# Harnessing LLMs as Agents: What Does It Cost?

**Zelin Zhao[1], Xinyu Guo[1], Jingyuan Zhang[1], Yuxuan Zhang[2], Yongxin Chen[1]**
[1]Georgia Institute of Technology, [2]Etude AI

Language-model agents increasingly rely on harnesses that manage bounded context, persistent memory, tools, verification, and repeated execution, but existing notions of model capability do not quantify the computational resources these mechanisms consume. We develop a theory of these costs through the *Language Model Agent Machine* (LAM), a resource-bounded model that fixes the underlying semantic model while explicitly charging harness-level resources. Our theory yields four classes of results. **Communication:** LAM execution and red–blue pebbling are instancewise equivalent at the level of simultaneous call–transfer budgets, yielding the full discrete resource frontier and transferring classical I/O lower bounds to context–memory traffic. **Access:** memory interface changes asymptotic cost, including a $\Theta(n)$ separation between random and non-speculative sequential access on pointer chasing and an exact single-stack FIFO frontier. **Recomputation:** bit-reversal DAGs require $\Theta(n^2/(C+S)+n)$ model calls with context capacity $C$ and persistent-memory capacity $S$, quantifying when stored intermediate state avoids repeated semantic computation. **Reliability:** we derive tight stage-local sampling bounds, exact imperfect-verification costs, and a Young–Daly-type checkpoint law with a closed-form optimal verification interval. Controlled experiments and held-out evaluations on GPT-6 Astra [1] compare development-derived communication and reliability predictions with measurements, including checkpoint optima and a held-out policy-selection tournament conditional on a programmatic segment check; held-out chained MATH additionally exposes an observed tradeoff between whole-task call granularity, logical input traffic, and reliability. Prescribed-schedule runs are reported separately as implementation checks. Together, these results provide a resource theory for the computational cost of language-model agent harnesses.

**Main Contact:** zelin@gatech.edu

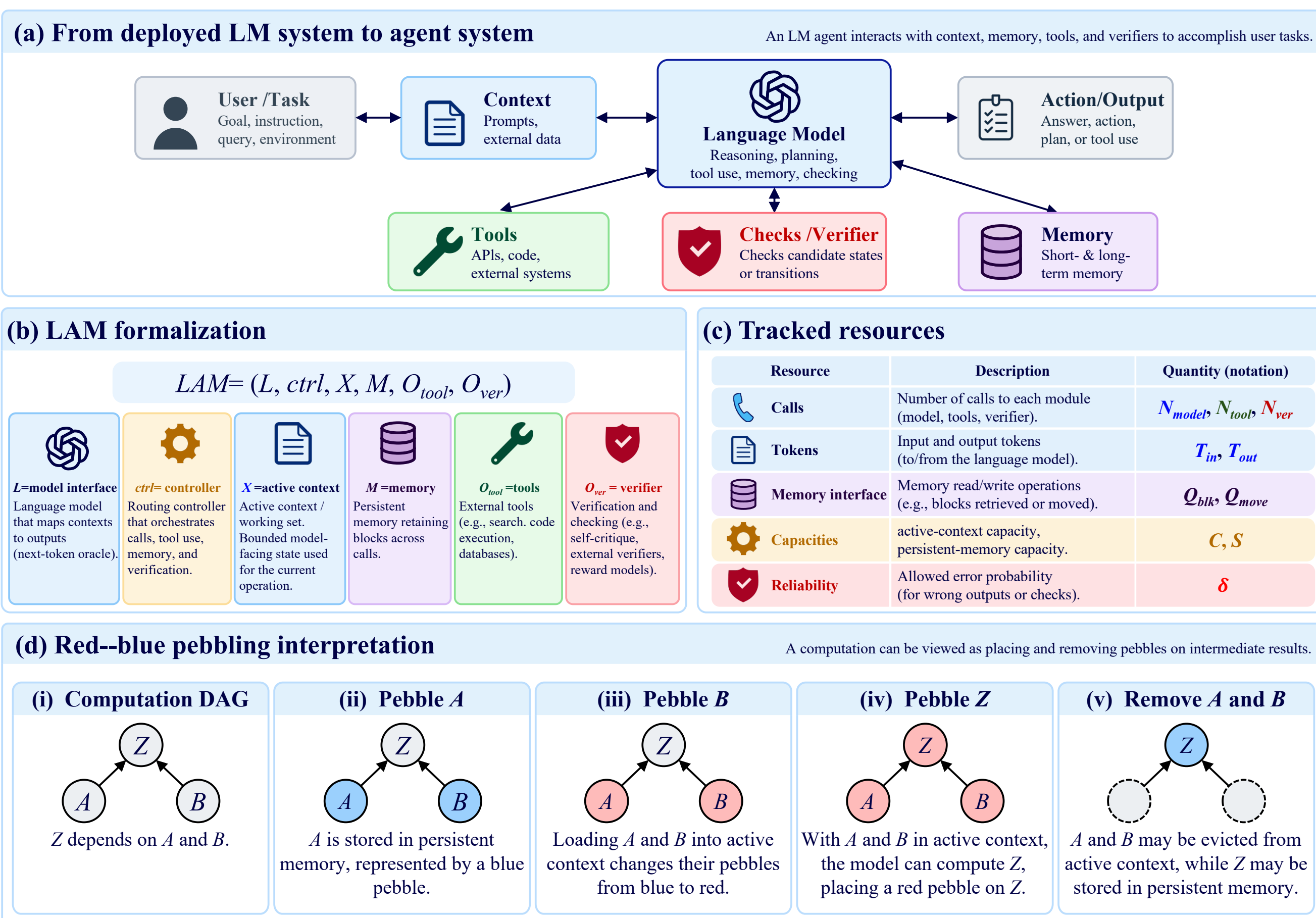

**Figure 1.** Overview of the LAM: (a) an agent with active context, persistent memory, tools and verifiers; (b) the LAM components; (c) the charged quantities (calls $N$, token traffic $T_{\text{in}}$, $T_{\text{out}}$, memory operations, capacities, target failure probability $\delta$); (d) red–blue pebbling (context blocks red, memory blocks blue, transfers communication).

## 1. Introduction

Modern language-model agents are computational systems built around a model, not isolated model calls: a harness manages context, persistent state, tools, verification, and repeated execution [2], [3], [4], [5]. Existing theory largely characterizes what models or agentic extensions can compute: additional reasoning steps, external memory, tool access, and verifier interaction can all change computational power [6], [7], [8], [9], [10]. What it does not characterize is the resource cost of realizing that computation through a harness.

Harness architecture is itself a source of computational complexity. Two agents using the same underlying model can have the same semantic capability yet require asymptotically different numbers of model calls, context–memory transfers, memory operations, or verification steps. We therefore ask a different theoretical question: holding semantic capability fixed, what resources must a harness spend to execute a computation, and what is the optimal tradeoff among model calls, context, persistent storage, communication, and verification?

We introduce the *Language Model Agent Machine* (LAM), a resource theory for model–harness execution. LAM fixes semantic-oracle power and makes context, persistent storage, state movement, model calls, tools, and verification explicit computational resources. Its value-provenance principle allows the harness to move, store, and reuse semantic state but prevents it from creating task-relevant content for free. This separation is essential: it lets us attribute differences in execution cost to harness design rather than to a stronger underlying model.

This viewpoint exposes resource questions that are invisible to computability alone. How much communication is forced by bounded context? Does the interface to the same amount of memory change complexity? When can persistent state replace repeated semantic computation? How expensive is correctness when model transitions are imperfect? Classical complexity theory provides important ingredients for each question, but harness executions introduce complications absent from the classical abstractions: values may be copied or recomputed, only active-context values can directly enter a model call, the controller must not store semantic information for free, and an incorrect model transition may be committed before its error can be detected. We resolve these complications in four resource laws:

- **Communication is exactly characterized by pebbling (Section 4).** Red–blue pebbling provides the classical abstraction for fast–slow memory communication, but a harness execution contains physical block copies, recomputation, and controller-managed state that do not correspond one-to-one with graph vertices. Using value provenance, we quotient block occurrences by computational origin and prove an instancewise, resource-preserving correspondence: for every $L$-realisable DAG, memory-initialized vertex-granular LAM evaluations and red–blue pebblings have exactly the same feasible call–transfer budgets. The result therefore preserves the full discrete Pareto frontier, not only an asymptotic I/O lower bound or a single weighted optimum (Theorem 1, Corollary 9 and Corollary 10). Classical pebbling lower bounds consequently become harness-communication lower bounds. On a sequential chain, dependency-only prompting uses $\Theta(nb)$ logical input symbols, versus $\Theta(n^2 b)$ for full-history replay (Proposition 12).
- **The memory interface changes computational complexity (Section 5).** Memory capacity alone does not determine harness cost. For FIFO retrieval through a single stack, arbitrary buffering and copying defeat the usual stack-order argument; we prove a single-copy normal form that removes this freedom without increasing context, and from it derive the exact context–storage–traffic frontier, including the precise feasibility region and optimal transfer cost (Theorem 17, Corollary 19). Random access solves the same routing task with constant active context. On pointer chasing, random access follows $k$ pointers in $k$ reads, whereas every always-correct non-speculative sequential execution requires $\Omega(kn)$ expected head movements; even unrestricted always-correct read-only sequential execution retains an $\Omega(n)$ lower bound (Theorem 2, Proposition 14).
- **Persistent state trades directly against semantic recomputation (Section 6).** The classical bit-reversal graph has a known call–space tradeoff, but the LAM setting separates active context from persistent storage: only context participates directly in a model call, while stored values must be explicitly recovered, and the input source is delivered only once. We lift the classical tradeoff to this asymmetric execution model and obtain a complete harness-level characterization: optimal model-call complexity $\Theta(n^2/(C+S)+n)$, explicit two-sided constants, the exact feasibility region, and a matching schedule with linear context–memory traffic (Theorem 3). In particular, linear model-call complexity is achievable if and only if total semantic capacity $C+S$ is linear in $n$; without persistence, the same computation incurs a $\Theta(n/C)$ multiplicative recomputation overhead when $C = o(n)$.
- **Correctness feedback changes the cost of reliable execution (Section 7).** Classical noisy-query bounds describe the cost of distinguishing a correct answer locally, but an agent commits to a sequence of dependent states. We construct a hidden-tree family in which an erroneous commitment produces no later diagnostic signal, allowing adaptive within-stage stopping while making mistakes irreversible. On this family, stage-local sampling without correctness feedback requires $\Theta(n\gamma^{-2}\log(n/\delta))$ model calls, and the order is tight (Theorem 21 and Theorem 4). Perfect local verification removes the logarithmic amplification overhead, while imperfect verification admits an exact accepted-error and retry-cost law (Theorem 5). Classical checkpointing supplies the square-root intuition, but under LAM we derive the exact rollback cost for semantic transitions and a closed-form optimal checkpoint interval $k^{\star}$ (Theorem 6).

**What the LAM abstraction adds.** The resource laws above do not follow from their classical counterparts by renaming memory or checkpoint variables. Each requires an execution-level bridge between a harness trace and the corresponding mathematical object. Table 1 summarizes the resulting theory.

The theory is also evaluated prospectively on GPT-6 Astra [1], a frontier reasoning model. Quantitative forecasts are determined from development or calibration data and evaluated on disjoint held-out data with the forecast quantities held fixed. Checkpoint-cost forecasts identify the observed optimal interval at every tested verification price, and a separately frozen tournament prospectively tests cost-aware checkpoint-policy selection, conditional on a programmatic segment check that is not priced as a semantic call. Held-out chained MATH confirms the input-traffic advantage of dependency-aware routing within stepwise execution while exposing an important boundary:

**Table 1. Harness-level content of the four resource laws.** Each row identifies the execution feature that must be handled to obtain an exact resource statement for a fixed semantic model.

| Resource | Harness-specific obstacle | LAM result |
|---|---|---|
| Communication | A harness may copy, evict, reload, and recompute physical block occurrences while the controller manages state outside the computation DAG. | Value provenance yields an instancewise correspondence with red–blue pebbling and equality of the feasible $(N, Q_{\text{blk}})$ budget region, hence the full discrete call–traffic Pareto frontier. |
| Access | Equal-capacity memories expose different admissible operations, and buffering or copying can invalidate ordinary stack or stream arguments. | Random versus non-speculative sequential access has a $\Theta(n)$ separation on pointer chasing, while FIFO retrieval admits an exact single-stack feasibility and traffic frontier. |
| Recomputation | Only active context participates directly in semantic calls, while persistent state must be explicitly recovered and the input may be delivered only once. | Bit-reversal execution has optimal call complexity $\Theta(n^2/(C+S)+n)$, with an exact feasibility region and a matching linear-transfer construction. |
| Reliability | Errors occur inside a dependent state trajectory and correctness feedback is itself a priced resource. | Stage-local sampling is tight, imperfect verification has an exact error–cost law, and rollback admits a closed-form optimal checkpoint interval. |

whole-task model calls can collapse many logical steps into a single call, with substantially lower input traffic but lower observed held-out reliability in our experiment. Prescribed schedules are reported separately as implementation checks rather than empirical tests of lower bounds (Section 8).

## 2. Related Work

**Agent computability and harness complexity.** Prior theory largely asks what language models and agentic extensions can compute, including the effects of additional reasoning, external memory, tools, and verification [2], [3], [4], [6], [7], [8], [10], [11]. Closest in spirit, MemGPT [12] manages a bounded context window by paging state to external storage, an operating-system analogy that our active context $X$ and persistent memory $M$ formalize. Differently, LAM asks a question: holding semantic-oracle power fixed, what resources must the harness spend to realize a computation? This makes model calls, context, persistent storage, memory traffic, and verification explicit computational resources, and exposes complexity differences between harnesses with the same semantic capability. Structured procedures handed to a model, such as program-guided transformers [13] or routines discovered from a single demonstration [14], structure-aware planning with a world model [15], or natural-language skills that a frozen model evolves from its own experience and verifier feedback [16], are harness-level objects in this sense: they change the execution schedule without changing the model. Research-level mathematics agents such as RMA [17], with persistent proof state, bounded per-call context, and verifier feedback, are concrete harnesses of the kind LAM is designed to account for.

**Communication, access, and recomputation.** Red–blue pebbling [18] and classical bit-reversal tradeoffs [19] provide lower-bound machinery for abstract computations. LAM establishes the bridge to harness execution: the pebbling correspondence preserves simultaneous call and transfer budgets and therefore the full discrete Pareto frontier, while the bit-reversal theorem gives the joint law $\Theta(n^2/(C+S)+n)$ for active context and persistent memory, including explicit constants and the exact feasibility region. Our access results further show that memory capacity alone is insufficient: the access interface itself can change asymptotic complexity. Search-augmented agents meet the same constraint empirically: uncontrolled retrieval can saturate the context, which motivates adaptive control of what evidence enters it [20]; LAM prices this context–memory traffic explicitly.

**Reliability and cost-aware execution.** Noisy-query theory, reliable computation, verifier-assisted generation, selective prediction [21], capacity-constrained review [22], and checkpointing study different aspects of computation under uncertainty [10], [23], [24], [25], [26], [27], [28].

LAM treats correctness feedback as a priced computational resource, yielding tight sampling bounds without feedback, exact error–cost laws for imperfect verification, and a closed-form checkpoint optimum under rollback. Empirical work measures deployment-specific accuracy–cost tradeoffs [29], [30] and evaluates agents beyond one-shot task success, for example with workflow-aware episodes [31] or teaching-over-solving diagnostics [32]; our theory instead predicts feasible resource tradeoffs and harness choices before execution, which we test prospectively in Section 8.

## 3. Language Model Agent Machines

The LAM is designed to separate two quantities that existing agent abstractions often conflate: semantic capability and the resources required to orchestrate that capability (Figure 1a). It models the harness-level execution around a fixed semantic interface—context construction, persistent state, memory access, tools, verification, and repeated calls—rather than the model's internal computation.

**Semantic block.** Fix an input size $n$ and a finite alphabet $\Sigma$. A *semantic block*, a fixed-width unit of task-relevant information, is an element of $\mathscr{B}_n := \Sigma^{b(n)}$, where $b(n)$ is the block width.

**Machine model.** A LAM is a tuple $\Lambda = (L, \text{ctrl}, X, M, O_{\text{tool}}, O_{\text{ver}})$, drawn in Figure 1b, where $L$ is the model interface, ctrl is the harness controller, $X$ is the active context, $M$ is the persistent memory, and $O_{\text{tool}}$ and $O_{\text{ver}}$ are the tool and verifier interfaces. The active context $X$ holds at most $C$ blocks available to the next model call; the persistent memory $M$ holds at most $S$ blocks across calls outside $X$, and a stored block must be retrieved into $X$ before a call can use it, the persistent copy staying in place. A call to $L$ consumes blocks in $X$ and inserts its output as a new block of $X$, so a call takes at most $C - 1$ argument blocks. A tool returns a semantic block through an explicitly charged operation. A verifier returns a tag (e.g. accept/reject in Section 7), inserts no block, and may inspect all $C$ resident blocks. Each of $L$, $O_{\text{tool}}$ and $O_{\text{ver}}$ is a fixed map of its opcode and argument blocks alone (the *interface clause*): no interface sees the input, $M$, or a block not passed to it. The controller is a finite control with tag and address registers that see block contents only through returned tags and decoded addresses, plus the occupancy pattern of $X$ and public step counters; Definition 3 (Appendix A) fixes the step rule and the operation set that every lower bound below quantifies over.

**Definition 1** (Value provenance)**.** A LAM execution satisfies *value provenance* (VPA) if every semantic block originates from an input, a declared constant, or a charged model or tool call. Copying, movement, and storage preserve this origin, while recomputation creates a new occurrence of the same computational value. The controller cannot create semantic blocks otherwise.

## 4. Communication: Moving State Across Bounded Context

**Communication cost.** For model-facing traffic, let $T_{\text{in}}$ be the total number of symbols supplied to model calls over an execution and $T_{\text{out}}$ the number produced by them; for context–memory traffic, let $Q_{\text{blk}}$ be the number of semantic blocks transferred between $X$ and $M$, that is $\Theta(Q_{\text{blk}} b(n))$ symbols.

**Computation graph.** We represent dependencies among semantic blocks by a DAG $G = (V, E)$ with one block per vertex. A *source* (no incoming edge) is an input block; an *internal vertex* is produced by a model call from the blocks of its predecessors, an edge $u \to v$ meaning that $u$'s block is required to compute $v$'s. $G$ is *$L$-realisable* if every internal vertex can be computed by the same interface $L$ from its predecessor blocks.

**Red–blue pebbling.** The red–blue pebble game [18] represents block movement between $X$ and $M$ (Figure 1d): a red pebble on $v$ means its block is in $X$, a blue pebble that a copy is in $M$, and a vertex may carry both. A `load` adds a red pebble to a blue-pebbled vertex, a `store` adds a blue pebble to a red-pebbled vertex, a `compute` places a red pebble on an internal vertex whose predecessors are all red, and either pebble may be `deleted`. Red pebbles are bounded by $C$, blue pebbles by $S$, and each load or store adds one to $Q_{\text{blk}}$. A pebbling starts with a blue pebble on every source and no red pebble, and ends with every designated sink (output vertex) red or blue; $\text{IO}_{C,S}(G)$ is the least number of loads and stores of such a pebbling ($\text{IO}_C(G)$ when $S$ is not binding). The analogy is not yet an equivalence: a LAM execution may copy and recompute semantic values, so physical block occurrences do not correspond one-to-one with DAG vertices. Value provenance lets us quotient these occurrences by computational origin while preserving both model-call and transfer budgets.

**Definition 2** (Vertex-granular $G$-evaluation)**.** A vertex-granular evaluation of an $L$-realisable DAG $G$ is a VPA-respecting LAM execution that produces the designated sinks of $G$, where each charged semantic call implements at most one declared vertex operation of $G$. The execution may copy, move, store, retrieve, evict, and recompute values arbitrarily. Under the *memory-initialized* convention, all source blocks are initially stored in $M$, the context is initially empty, and every designated sink is resident in $X$ or $M$ at termination. Under the *input-delivered* convention, each source is instead delivered exactly once by INPUT.

**Theorem 1** (Red–blue pebbling correspondence)**.** *Let $G$ be an $L$-realisable computation DAG with designated sinks. Every memory-initialized vertex-granular $G$-evaluation with capacities $(C, S)$ induces a legal red–blue pebbling with at most $C$ red and $S$ blue pebbles, exactly one* COMPUTE *move per model call, and at most one* LOAD/STORE *move per context–memory transfer. Conversely, every such pebbling induces a* *vertex-granular evaluation on a* RAM-LAM *with no larger call or transfer budget. Hence the two models have the same feasible call–transfer budget pairs and the same discrete Pareto frontier.*

The theorem identifies the full resource region, not merely an individual classical I/O lower bound. Every red–blue pebbling lower bound therefore becomes a harness-communication lower bound; weighted cost objectives follow immediately (Corollary 9). For example, the classical $n \times n$ grid bound yields $Q_{\text{blk}} = \Theta(n^2/C)$ for $3 \le C \le n$ when persistent storage is nonbinding (Corollary 11).

## 5. Access: Retrieving Information from External State

**Memory-access interfaces.** A *random-access* interface reads a block directly from a specified address; a *sequential* interface stores blocks in an ordered stream accessed through a movable head. Let $Q_{\text{move}}$ be the number of unit head movements and $Q_{\text{mem}} \coloneqq Q_{\text{blk}} + Q_{\text{move}}$ the equal-weight memory-interface operation count (written $Q$ in the appendices); for random access $Q_{\text{move}} = 0$ (separate deployment weights in Appendix E).

**Pointer chasing.** A memory table has $n$ entries, entry $i$ storing an index $\pi(i) \in [n]$ for a permutation $\pi$ of $[n]$. Starting from $a_0$, the agent reads entry $a_j$ to obtain $a_{j+1} = \pi(a_j)$ for $k$ steps and outputs $a_k$. One semantic block holds one index in $[n]$, so $b(n) = \lceil \log_{|\Sigma|} n \rceil$, and $a_0 = 1$ is fixed. For the sequential interface the table is read-only, entry $i$ occupies cell $i$ and the head starts at cell 1; we consider *non-speculative* executions: every READ is of cell $a_0$ or of a cell whose index is the content of a block previously read into $X$.

**Theorem 2** (Demand-following pointer chasing). *For the pointer-chasing task above, let $\pi$ be uniformly random over permutations of $[n]$, and let $2 \le k \le n/4$. A random-access* LAM *follows the $k$ pointers with $Q_{\text{blk}} = k$ and $Q_{\text{move}} = 0$: at $C = 1$ with no model call when* SETADDR *decodes each entry, at $C = 2$ with $k$ calls when an oracle call decodes each entry instead. Every always-correct, non-speculative sequential* LAM*, deterministic or zero-error randomised, satisfies $\mathbb{E}[Q_{\text{move}}] \ge \frac{3}{256} kn$ over $\pi$ and its random tape, hence $\mathbb{E}[Q_{\text{mem}}] = \Omega(kn)$.*

*Proof idea.* Non-speculation forces first reads to follow the pointer orbit, whose next address is uniform over the unvisited cells, so each of the $k \le n/4$ hops moves the head $\Omega(n)$ in expectation (Appendix C.1). Beyond this class, Proposition 14 gives $\Omega(n)$ for every always-correct read-only sequential execution, and Remark 3 a speculative bounded-workspace machine with $O(\min\{kn, (n^2/C) \log C\})$ expected head movements for $C \ge 12$.

**Exact one-stack frontier (Corollary 19).** In $\text{FIFO}_n$, blocks arrive in order during a write phase and must be emitted in order after all arrivals. Consider a one-STACK LAM with no model or tool call, tag-only verifiers, arbitrary finite control and $|\mathscr{B}| \ge n + |K_0|$ for its constant set $K_0$ (Theorem 17). Its exact worst-case frontier is

$$\text{feasible} \Leftrightarrow C \ge \lceil (n+1)/2 \rceil \text{ and } S \ge (n-C)_+, \qquad Q^\star_{\text{blk}} = 2(n-C)_+,$$

with lower bounds valid under arbitrary write-phase buffering, whereas copy-free streaming requires $C \ge n$ (Theorem 17(iii)); a content-blind schedule attains the frontier (Corollary 19, Appendix C.3), and random access works at $C = 1$, $S \ge n$. *Proof idea.* At the last push of the bottom block every block of smaller index is resident, at its first pop every block of larger index, and the bottom block at both instants, so $n + 1 \le 2C$; a single-copy normal form makes the count independent of buffering (Lemma 18, Appendix C.3).

## 6. Recomputation: Persistence versus Regeneration

**Persistence and recomputation.** In the bit-reversal DAG many intermediate results are reused after leaving active context. Let $n = 2^m$ and let $\text{rev}\,(i)$ reverse the $m$-bit binary string of $i \in \{0, \dots, n-1\}$. The DAG $\Delta_n^{\text{rev}}$ consists of two length-$n$ chains $u_0, \dots, u_{n-1}$ and $w_0, \dots, w_{n-1}$ with edges $u_i \to w_{\text{rev}\,(i)}$, so the first chain's results are reused by the second in a different order.

**Theorem 3** (Persistence versus recomputation). *Let $\Delta_n^{\text{rev}}$ be $L$-realisable and $n$ a power of two. Over input-delivered vertex-granular evaluations of $\Delta_n^{\text{rev}}$ (Definition 2) with context $C \ge 3$, random-access persistent memory of capacity $S \ge 0$ and $C + S \ge 4$, the least number of model calls is $N = \Theta(n^2/(C+S) + n)$, explicitly $\frac{1}{32}(\frac{n^2}{C+S} + n) \le N \le 4(\frac{n^2}{C+S} + n)$, the upper bound attained by a* RAM-LAM *with $Q_{\text{blk}} \le 2n$; the displayed lower bound also holds for every input-delivered vertex-granular $G$-evaluation with $C \ge 2$ and $S \ge 0$. For $n \ge 4$, an input-delivered vertex-granular $G$-evaluation exists if and only if $C \ge 3$ and $C + S \ge 4$. Thus $S \ge n$ gives $\Theta(n)$ calls with $Q_{\text{blk}} = O(n)$, under the memory-initialized convention as well, and $S = 0$ requires $\Theta(n^2/C + n)$.*

Within the feasible region, $N = \Theta(n)$ if and only if $C + S = \Omega(n)$: sublinear total semantic capacity forces superlinear calls however it is split between context and persistence, the split setting the transfer cost (Appendix C; Appendix G.3). *Proof idea.* For $C + S \le n/2$, the $u$-arguments of $L = \Theta(C+S)$ consecutive outputs lie $n/L$ apart and at most $C + S$ of the gaps hold a pebble, so each window forces $n/4$ fresh first-row placements (Lemma 20); $n/L$ disjoint windows give the $n^2/(C+S)$ term, and evenly spaced checkpoints attain it.

## 7. Reliability: Verification and Checkpointing

Consider a chain of $n$ semantic transitions in which, given a correct state, a model call returns the intended successor with probability $1-\varepsilon$; an undetected error propagates to the end; we ask how sampling and verification change the cost of an overall failure probability at most $\delta$.

**Sampling without correctness feedback.** For independent repeated calls at a fixed state, let $\gamma > 0$ denote the probability gap between the correct successor and the most likely incorrect alternative. We call a strategy *stage-local sampling-only* if it may draw an adaptive number of samples at each step and aggregate them arbitrarily, but receives no correctness signal and cannot revise a committed transition after moving to the next step. Plurality implements self-consistency [26]: under independent samples, if at most $A$ incorrect alternatives have nonzero probability at each correct state, each with probability at most $p_c - \gamma$, a plurality vote over $m = O(\gamma^{-2}\log(An/\delta))$ samples per transition reaches overall failure probability at most $\delta$ with $N = O(n\gamma^{-2}\log(An/\delta))$ model calls (Theorem 21, Appendix D); on the binary hard family ($A = 1$) this order is optimal, so the upper and lower orders match for bounded $A$.

**Theorem 4** (Stage-local sampling lower bound)**.** *There exists a depth-$n$ binary-tree family of $\gamma$-separated transition distributions, $\gamma \leq 1/2$, its nodes encoded as blocks of width $b(n) = \Theta(n)$, with hidden labels drawn at random, such that every stage-local sampling-only strategy that knows the construction but not the labels, samples only at its current node, and reaches the designated leaf with probability at least $1-\delta$, $\delta \leq 1/4$, over labels, samples and its own randomness, satisfies $\mathbb{E}[N] \geq \frac{n}{4\gamma^2}\log\frac{n}{8\delta}$, which is $\Omega(n\gamma^{-2}\log(n/\delta))$ for $n \geq 64\delta$.*

Off-path states have the same local sampling law as correct-path states, so an error never reveals itself through later outputs (proofs in Appendix D).

**Local verification.** With verification the agent receives a correctness signal before committing to the next state. Let a verifier call have weighted cost $c_v$ relative to one model call, so that $\text{Cost} = N_{\text{model}} + c_v N_{\text{ver}}$. With a perfect local verifier, fresh independent attempts of error probability $\varepsilon < 1$, and uncapped retries, retry-until-accept terminates almost surely with zero error and expected weighted cost $\mathbb{E}[\text{Cost}] = n(1+c_v)/(1-\varepsilon)$ (Appendix D, Theorem 22). On the hard family of Theorem 4 this removes the $\Theta(\log(n/\delta))$ factor for constant $\gamma$ and $c_v$: the gain comes from feedback, not from more sampling. For an implemented checking procedure with joint acceptance masses $a_t = \Pr(\text{correct}, A_t)$ and $b_t = \Pr(\text{wrong}, A_t)$, i.i.d. attempts give accepted-error probability $q_t = b_t/(a_t+b_t)$ and mean attempt count $1/(a_t+b_t)$, dependent checks within an attempt included; Theorem 5 below is the content-independent, conditionally independent case, generalised in Corollary 23–Corollary 24 (Appendix D).

**Theorem 5** (Imperfect local verification)**.** *Let $p_a$ be the probability that the verifier accepts a wrong candidate and $p_r < 1$ the probability that it rejects a right one, the same whatever the candidate's content, and let $0 < \varepsilon < 1$. Suppose candidate attempts are independent, the $t \geq 1$ checks of a candidate are conditionally independent given its correctness, every attempt pays for all $t$ checks, and a candidate is accepted only if all $t$ accept. Define $a_t = (1-\varepsilon)(1-p_r)^t$, $b_t = \varepsilon p_a^t$, and $s_t = a_t + b_t$. Then, at any transition whose earlier accepted transitions are all correct, the accepted candidate is incorrect with probability $q_t = b_t/s_t$ and the expected numbers of attempts, model calls and verifier calls are $1/s_t$, $1/s_t$ and $t/s_t$; hence $\Pr[\text{chain failure}] \leq nq_t$. If the attempt law is the same at every visited state, the chain fails with probability exactly $1-(1-q_t)^n$, and summing over the $n$ transitions gives the unconditional totals $\mathbb{E}[N_{\text{model}}] = n/s_t$, $\mathbb{E}[N_{\text{ver}}] = nt/s_t$ and $\mathbb{E}[\text{Cost}] = n(1+tc_v)/s_t$.*

The accepted-error odds strictly decrease with $t$ if and only if $0 < p_a < 1-p_r$, since $q_t/(1-q_t) = \frac{\varepsilon}{1-\varepsilon}\left(\frac{p_a}{1-p_r}\right)^t$; at $p_a = 0$ one check makes them zero. Under the stationary attempt law at fixed $(\varepsilon, p_a, p_r)$ with $0 < p_a < 1-p_r < 1$ and $n\varepsilon > \delta$, the smallest $t$ with $nq_t \leq \delta$ gives $\mathbb{E}[N_{\text{model}}] = \Theta(n(n/\delta)^{\alpha})$, $\alpha = -\log(1-p_r)/\log((1-p_r)/p_a)$, with $t = \Theta(\log(n/\delta))$ (Corollary 25; varying-$\varepsilon$ bracket, capped-retry law and short-circuit costs in Appendix D, Remark 8).

**Checkpointing.** Alternatively, verify every $k$ transitions and roll a rejected segment back to the last accepted checkpoint.

**Theorem 6** (Checkpoint law)**.** *Assume independent transition errors with probability $\varepsilon$, a perfect segment verifier of cost $c_v$, and a policy that verifies every $k$ transitions and rolls back the entire rejected segment. If $k$ divides $n$, its expected weighted cost is $T(k) = \frac{n}{k}(k+c_v)(1-\varepsilon)^{-k}$. For $0 < \varepsilon < 1$ and $c_v > 0$, writing $\beta = -\log(1-\varepsilon)$, the positive-real minimizer is $k^\star = \frac{1}{2}\left(\sqrt{c_v^2 + 4c_v/\beta} - c_v\right)$. When $c_v\beta \to 0$, $k^\star = \sqrt{c_v/\beta}(1 + O(\sqrt{c_v\beta}))$, and $k^\star \sim \sqrt{c_v/\varepsilon}$ if also $\varepsilon \to 0$.*

The square-root scaling is the checkpointing law of J. W. Young [27], [28] in the whole-segment form of silent-error checkpointing (Appendix D); an optimum of the integer extension of $T(k)$ lies within one of $k^\star$, and a verifier cost linear in $k$ returns the optimum to $k = 1$ (Appendix D). For absorbing-error length-$k$ rollback attempts costing $k + c_s$ under a perfect segment verifier, conditional hazards in $[\underline{\varepsilon}, \overline{\varepsilon}]$, $\overline{\varepsilon} < 1$, after every history and rollback bracket $\mathbb{E}[T(k)]$, $k \mid n$, between $T(k)$ at $\underline{\varepsilon}$ and at $\overline{\varepsilon}$, without independence within or between attempts (Proposition 28, Appendix E).

**A joint decision.** Local checks and rollback combine into a policy cheaper than either mechanism alone in an explicit regime; Proposition 26 (Appendix E) prices it exactly, $T(t,k) = \frac{n}{k}(kg_t + c_s)(1-q_t)^{-k}$, with $g_t = (1+tc_v)/s_t$ the expected weighted cost per locally accepted step, $q_t = b_t/s_t$ its residual error (Theorem 5) and $c_s$ the perfect segment-verifier cost (transfer price in Corollary 27).

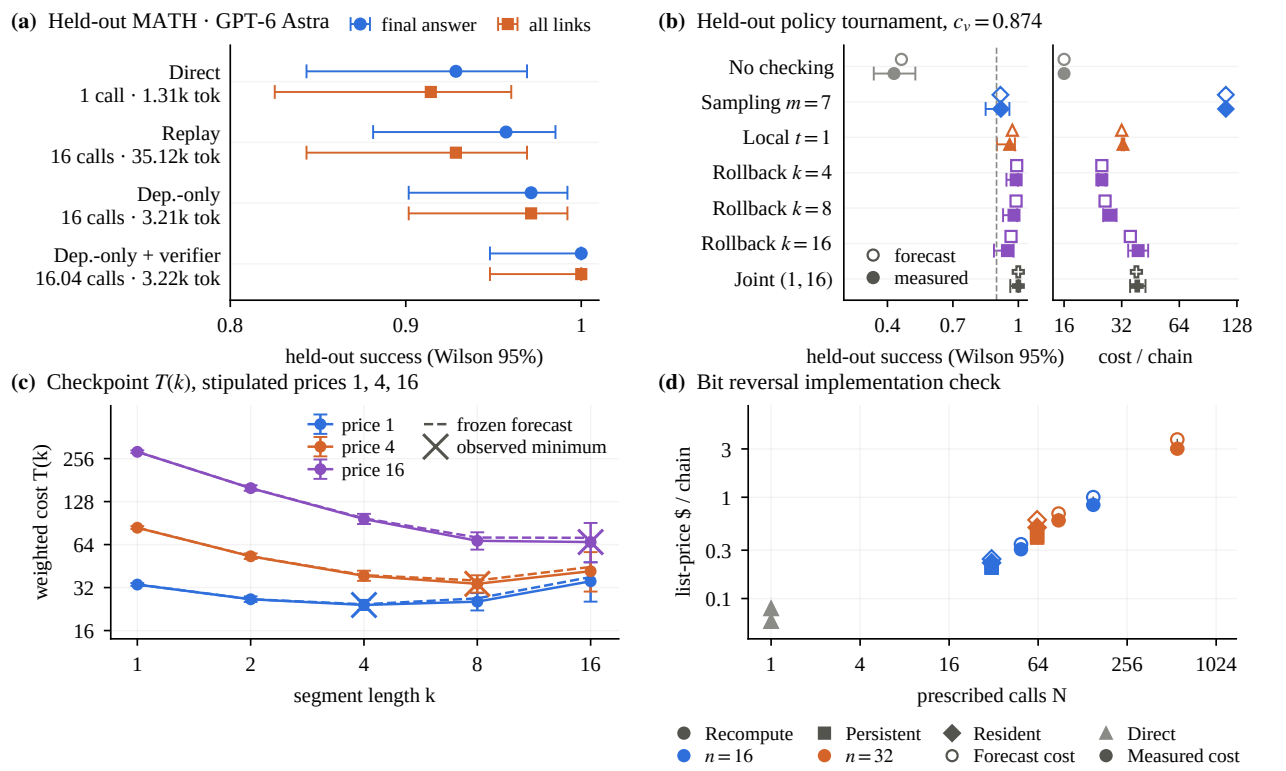

**Figure 2. GPT-6 Astra evaluation: prospective predictions, held-out outcomes, and implementation checks.** All model calls in panels (a)–(d) use the same fixed GPT-6 Astra deployment; panel (c) uses a programmatic perfect segment verifier at stipulated prices. Where predictions are shown, hollow markers and dashed lines denote development- or calibration-derived predictions, while filled markers and solid lines denote measurements with 95% intervals. **(a)** Held-out MATH final-answer and all-link success on 70 chains per condition. **(b)** Held-out checkpoint-policy tournament on 100 fresh chains per arm at measured verifier price $c_v = 0.874$; the horizontal dashed line is the target success probability 0.90. **(c)** Checkpoint cost $T(k)$ under stipulated perfect-verifier prices $1, 4, 16$; forecast and measured minimisers coincide at each price. **(d)** Bit-reversal implementation check under enforced payload limits; marker shape denotes schedule type, color denotes problem size $n$, and hollow/filled markers denote forecast/measured cost.

# 8. Experiments

Although **our primary contribution is theoretical**, we use experiments in two roles. All model-based experiments use the same fixed GPT-6 Astra deployment. *Held-out tests* compare development-derived quantitative predictions or policy choices with disjoint evaluation data, including communication, checkpointing, and reliability. *Implementation checks* verify that prescribed schedules realize their analytically determined resource accounting; these checks are reported separately and are not treated as empirical tests of lower bounds. Protocols and evaluation details appear in Appendix G.

**Communication.** As Proposition 12 predicts, replay-to-dependency-only logical input grows linearly with depth: $7.1\times$ at $n = 64$ on the synthetic chain ($1.72\times$ under the exact-prefix cache model) and $11\times$ ($3.05\times$ cached) on development MATH (Appendix G.1).

## 8.1. Held-out MATH: traffic, success, and granularity

From the 207 eligible MATH-500 problems [33], [34] we build 100 depth-16 answer-dependent chains following Compositional GSM [35]: chains 1–30 form the development set and 31–100 the held-out compositions. We evaluate GPT-6 Astra under four harness configurations: Direct, using one whole-chain call; full-history replay; dependency-only prompting; and dependency-only prompting with an ACCEPT/REJECT verifier. The routing protocol, traffic intervals, and success forecasts are determined from the development chains and held fixed on the held-out compositions.

On the held-out compositions, dependency-only prompting with verification solves 70/70 chains with every link correct, whereas Direct solves 65/70 final answers and has 64/70 chains with every link correct. Replay solves 67/70 final answers and has 65/70 chains with every link correct (Figure 2(a)). Replay uses 35,124 input tokens per chain, versus 3,217 for dependency-only prompting with verification, a $10.9\times$ reduction, while Direct uses only 1,311 input tokens in a single whole-chain call. Thus dependency-aware routing sharply reduces logical input traffic within stepwise execution, while Direct exposes a complementary granularity tradeoff: one model call can realize many logical steps with still lower input traffic, but here with lower observed held-out reliability. The vertex-granular theory therefore characterizes the resource frontier of stepwise execution rather than unrestricted whole-task calls.

**Execution granularity is itself a resource choice.** The Direct condition highlights a distinction that is not visible in a vertex-granular cost model. A sufficiently capable model may compress many logical dependencies into one semantic call, simultaneously reducing both call count and input traffic. This does not contradict the vertex-granular frontier: it changes the computational primitive being priced. Conversely, the stepwise conditions expose intermediate state to memory management and verification, yielding higher all-link reliability in the held-out GPT-6 Astra evaluation. Thus the choice of semantic-call granularity determines which communication–reliability tradeoffs are available to the harness.

## 8.2. Reliability: checkpoint forecasts and held-out policy selection

An arm is a policy of Section 7 on a length-16 modular-affine chain $x \leftarrow (a\,x + b) \bmod 10^{16}$ with target failure probability $\delta = 0.10$. Forecast quantities are fixed before evaluation and compared against held-out measurements and their 95% intervals. The deployment's calibrated step error is $\hat{\varepsilon} = 0.049$, and its YES/NO check costs $c_v = 0.874$ solver calls in tokens (Appendix G.2). On a separate transfer set, unamplified chain success is forecast at 0.448 and measured at 51/100, while the in-chain step hazard is forecast and measured at 0.0489.

As a controlled check of the checkpoint law, we use a programmatic perfect segment verifier at stipulated prices $c_v^{\text{seg}} \in \{1, 4, 16\}$. Checkpoint attempts per segment match the frozen predictions at $k = 1/4/16$ (1.051/1.222/2.230 forecast versus 1.049/1.208/2.083 measured), the $k = 4$ cost forecast 24.4 is measured at 24.2 [22.3, 26.3], and forecast and measured grid minimisers coincide at every price ($k = 4/8/16$; Figure 2(c)).

Wrong-sample rates vary sharply across chain states, so a pooled homogeneous i.i.d. error model does not describe the observed sampling data.

**Held-out policy tournament.** Because selection from measured comparisons can be sensitive to perturbations of the comparison data [36], a decision rule calibrated on disjoint data is held fixed on 100 shared held-out chains per arm. At the primary setting $(\delta, c_v) = (0.10, 0.874)$, the frozen decision rule selects rollback $k = 4$. On the held-out chains, $k = 4$ is also the cheapest certified policy, with $1/100$ failures and weighted cost 25.2 [24.1, 26.4], matching the frozen cost forecast of 25.2.

The tournament uses an implemented segment-check interface: public affine constants are composed into $x \mapsto (Ax + B) \bmod 10^{16}$ outside the priced semantic execution. We therefore treat the result as a prospective test of checkpoint-policy selection conditional on that interface, rather than as an end-to-end test of the VPA restriction (Appendix G.2.0.3).

**Implementation check.** The prescribed bit-reversal schedules reproduce their analytically determined call, recomputation, and transfer counts exactly. Figure 2(d) places these schedules at their prescribed call counts and compares frozen with measured list-price cost; we treat the count identities as implementation checks rather than empirical tests of the lower bounds (Appendix G.3).

## 9. Conclusion

LAM makes agent-harness architecture a mathematical object whose resource requirements can be bounded, compared, and optimized while semantic capability is held fixed. Across communication, memory access, recomputation, and reliability, the framework yields exact feasible regions, lower bounds, and closed-form resource tradeoffs. The experiments compare frozen predictions with measurements while also exposing the scope of the theory: dependency-aware routing sharply reduces logical input traffic within stepwise execution, whereas held-out chained MATH shows that a whole-task call can use still less input traffic at lower observed reliability, outside the vertex-granular frontier. Together, these results separate semantic capability from the resource cost of realizing that capability through an agent harness. LAM therefore provides a resource theory of agent execution: harness architecture determines which call, memory, communication, and reliability tradeoffs are achievable even when the underlying semantic model is held fixed.

## AI Use Statement

The research questions, conceptual framework, theoretical formulation, experiments, and final conclusions in this work were developed and evaluated by the authors. The authors made all substantive research decisions, verified the retained mathematical claims and proofs, and take full responsibility for the final content of the paper.

AI assistance was used to aid or polish writing, for retrieval and discovery such as finding related work, and for proving mathematical claims, including proof search, proof checking, and searching for logical gaps, missing assumptions, or counterexamples. AI outputs were treated as suggestions rather than evidence of correctness and were independently reviewed by the authors.

## Reproducibility statement

All theoretical results in the paper are stated with their assumptions and proved in the appendices. We release the simulation code, frontier-model harness, theorem-target problem set, figure-generation scripts, raw request logs, frozen experimental configurations, and result files used in the paper. The supplementary material also records the experimental protocols, prompts, model settings, datasets, seeds, and other parameters needed to reproduce the reported experiments.

### Ethics Statement

This work develops a theoretical framework for analysing the resource requirements of language-model agents and their harnesses. It does not involve human-subject research or the collection of personally identifiable or sensitive data. The empirical components are limited to controlled simulations and model-based experiments on a synthetic task and on public benchmark problems, intended to test the accounting and the assumptions of the theoretical model rather than to evaluate or deploy an agent in a high-stakes setting.

The framework may inform the design of more efficient and reliable agentic systems. As with general improvements in language-model agents, such advances could have both beneficial and potentially harmful downstream applications. Our work does not introduce methods specifically intended to facilitate harmful behavior, bypass safety mechanisms, compromise privacy or security, or enable autonomous operation in high-stakes domains. Moreover, many of the results rely on explicitly stated abstractions and assumptions and should not be interpreted as guarantees about the safety or reliability of deployed agents.

**Table 2.** Prior models of language-model computation as LAM parameter settings. $T$ is the number of reasoning steps. Context $C$ is measured in blocks; physical symbol/token width is $w = Cb$. The rows identify which harness resources are permitted to scale; computational power may also differ across settings, as illustrated by Proposition 7 and Proposition 8.

| Model | Memory | Context $C$ | Verifier | Resource that grows |
|---|---|---|---|---|
| Direct inference | — | $O(1)$ | — | nothing ($N = 1$) |
| Chain of thought [3] | — | $\Theta(T)$ | — | context |
| Managed context [37] | — | $O(1)$ | — | calls only |
| Memory-augmented [8] | ram | $O(1)$ | — | calls, memory I/O |
| Tool-augmented [11] | any | $O(1)$ | — | oracle queries |
| Verifier-assisted [10] | any | $O(1)$ | yes | verifier queries |
| **LAM (this work)** | any | any | optional | all of them, jointly |

## A. Formal machine semantics

**Relation to prior inference abstractions.** Table 2 places several existing inference abstractions within the resource coordinates of LAM.

**Configurations and steps.** The following definition fixes the machine semantics used by the lower bounds and constructive schedules throughout the paper.

**Definition 3** (LAM configuration and step)**.** Fix $n$, the block space $\mathscr{B} = \Sigma^{b(n)}$, a finite tag alphabet $\Gamma$, a finite set $K_0 \subset \mathscr{B}$ of constant blocks, and capacities $C$ and $S$. A configuration consists of

1. a finite-control state $q \in Q_{\text{ctrl}}$, $Q_{\text{ctrl}}$ finite;
2. the active context $X$: $C$ slots, each empty or holding one block. Every place in which the harness keeps a block outside memory, including any named block register, is one of these slots;
3. the persistent memory $M$, an interface-specific store of at most $S$ blocks ($S$ may be unbounded): for RAM an array of cells $M[1], \ldots, M[S]$; for STREAM a tape of cells with a head position; for STACK a sequence accessible at its top only;
4. two registers: a tag register holding the last tag $\gamma \in \Gamma$ and, for RAM, an address register $a \in [S]$. A STREAM machine has no address register: its only address state is the head position $a \in [S]$, which the controller may consult but which changes only by MOVE($\pm 1$); a STACK machine has neither;
5. public structural counters: finitely many integers whose values after $t$ steps are fixed functions of $t$ and $n$ alone (a step counter and fixed functions of it); they can never depend on the input.

Initially the context is empty, the registers hold fixed values, and $M$ holds the input for tasks whose input is held in memory. Every block carries an *occurrence label*: the input coordinate it copies, the constant it is, or the call occurrence that produced it; copies, reads and writes inherit the label. In a vertex-granular $G$-evaluation (Definition 2), each value-producing call is assigned a vertex of $G$; its vertex label is that vertex, while recomputation creates a new occurrence with the same vertex label. "Provenance label" means the occurrence label unless a vertex is named. An *input block* is a block labelled by an input coordinate. The registers $(q, \gamma, a)$ hold $h := \log_2 |Q_{\text{ctrl}}| + \log_2 |\Gamma| + \log_2 \max\{S, 1\}$ bits, the last term being the address or head state for RAM and STREAM (zero when $S = 0$) and omitted for STACK. Block contents are never visible to the controller except through the operations below that return a tag or set the address register. A step applies the controller transition function $\tau_{\text{ctrl}}$ to $(q, \gamma, a)$, the counters and the occupancy pattern of $X$ (which slots are full, not what they hold) to choose the next control state and one operation:

- CALL($L$, opcode, slots), and likewise a tool call: the blocks in the named occupied slots, at most $C - 1$ of them, are the arguments; the returned block is placed in a free slot (the insertion convention) and the returned tag in the tag register; the arguments are unchanged. A verifier call, CALL($O_{\text{ver}}$, opcode, slots), returns a tag only and inserts no block, so the $C - 1$ cap does not apply to it: its arguments may be all $C$ resident blocks (Remark 9 uses two at $C = 2$). *Interface clause:* $L$, each tool and each verifier is a fixed map from (opcode, argument-block tuple) to (block, tag), or to a tag alone for a verifier; it has no access to the input, to $M$, to the registers or to any block not among its arguments, so equal opcodes and arguments always return equal results.
- COPY(slot → free slot), EVICT(slot), and CONST($k$ → free slot) for $k \in K_0$.
- Memory operations. RAM: READ($a$) copies $M[a]$ into a free slot and WRITE($a$, slot) copies the slot into $M[a]$, the address being the address register or a fixed function of the public counters. STREAM: MOVE($\pm 1$), counted by $Q_{\text{move}}$ and the only operation that changes the head position, and READ/WRITE at the head. STACK: PUSH(slot) moves the block from the slot to the top of the stack and POP moves the top block into a free slot. Reads, writes, pushes and pops are counted by $Q_{\text{blk}}$; a READ of a cell that holds no block delivers the blank constant $\perp \in K_0$, with constant provenance.
- SETADDR(slot) and SHIFT($\pm 1$), for RAM only: the address register takes the value encoded by the block in the slot (this is how a decoded pointer becomes an address), or changes by $\pm 1$, clipped to $[1, S]$, reading no block. No interface turns block content into a position for free: a STREAM head reaches a cell only through MOVES charged to $Q_{\text{move}}$, one per cell traversed, and a stack only through PUSH/POP charged to $Q_{\text{blk}}$; Theorem 2 depends on this.

- INPUT: the next input block is delivered into a free slot, for tasks whose input arrives as a stream; for the pebbling results the input is held in $M$ instead.
- EMIT(slot): the block in the slot is the next output block; a halting step may also emit a tag.

The *observation counter* $Q_{\text{obs}}$ is charged one unit by every event that exposes an input block outside a model-call argument list: a SETADDR on an input block, each input block among the arguments of a tool or verifier call (per block, not per call), and each EMIT of an input block. Model-call arguments are not charged to $Q_{\text{obs}}$; they are counted through $N$. A randomised LAM additionally gives $\tau_{\text{ctrl}}$ a random tape. *Content-dependent state.* Besides the registers, $\tau_{\text{ctrl}}$ reads the occupancy pattern and the counters, and both are functions of the controller's past choices, hence content-dependent through the tags: the pattern carries up to $C$ bits, and a step counter carries $\log_2 t$ bits, which a machine can write by waiting, i.e. by performing a content-dependent number of steps that move no block. The controller-visible content-dependent state of a machine is therefore the registers, the pattern and the counters, $h + C + O(\log t)$ bits in all; the context adds $C\, b(n) \log_2 |\, \Sigma \,|$ bits of block content, which a later call can re-expose through its tag, and a writable memory adds its contents and their arrangement. No lower bound of Section 4–Section 6 bounds this state (the simulation of Theorem 16, Appendix C.2, carries all of it; Theorem 2 and Theorem 17 are pointwise); Remark 3, the one place that does, counts every part. Value provenance is a consequence of the list: INPUT, CONST and the returning calls are the only operations that create a block, and every other operation copies or moves an existing one. Every lower bound of Section 4–Section 6 quantifies over all machines of this definition with the interface, capacities and call restrictions named in its statement, and every upper bound exhibits one such machine; the reliability results of Section 7 are stated over the strategy classes defined there, whose stochastic oracle is an interface that also consumes fresh bits of the random tape at each call, the interface clause holding conditionally on those bits (Remark 9 runs the checkpoint policy as a machine of this definition). A machine of this definition is fixed for one $n$: the lower bounds are instancewise and therefore hold against non-uniform families, and every upper bound given as a schedule is a uniform family whose finite control, interfaces and $K_0$ do not depend on $n$, which enters only through the public counters, the one exception being the completeness direction of Theorem 1(ii), which builds one machine per pebbling; Proposition 7 concerns a single machine reading inputs of every length.

**Proposition 7** (Finite-state baseline)**.** *Let $\Lambda$ be a single deterministic* LAM *of Definition 3, its finite control, interfaces, tag alphabet and constants fixed independently of the input length, with $S = 0$ (no persistent memory), fixed block width $b$, total active semantic capacity $C$ blocks, and no public structural schedule counters, reading its input as a stream of blocks. Then the language decided by $\Lambda$ is regular; conversely every regular language is decided by such a* LAM *with $C = 2$. The one-pass input stream is an external input interface: after the current symbol is consumed it is not persistent memory and cannot be revisited.*

*Proof.* A configuration of $\Lambda$ between input blocks is, by Definition 3, the tuple (control state, contents of the at-most-$C$ active semantic slots, last tag, address register). There are at most $|\, Q_{\text{ctrl}} \,| \cdot (|\, \mathscr{B} \,| + 1)^C \cdot |\, \Gamma \,| \cdot (S + 1)$ such configurations, a finite number because $b$ and hence $\mathscr{B} = \Sigma^b$ are fixed in this proposition and $S = 0$. Named routing block registers do not contribute an additional factor: by definition they are names for slots inside the same $C$-block active semantic store. Public structural counters are excluded in this finite-state sanity check; they are used only in the explicit uniform schedule constructions stated elsewhere in the paper.

Because the oracle is a fixed map of its arguments (interface clause) and the controller's transitions depend only on the configuration and the next input block, the map (configuration, input block) $\mapsto$ configuration is a well-defined finite transition function, i.e. a DFA over the alphabet $\mathscr{B}$. Conversely, given a DFA $D$, encode the DFA state in the controller's finite control, pass the current input block to an oracle implementing $\delta_D$, and return the next DFA state as a constant-size tag. The input and returned block co-reside, so $C = 2$ under our insertion convention. □

**Proposition 8** (Resource-accounted universality)**.** *For every one-tape Turing machine $T$ running in time $t(n)$ and space $s(n)$ there is a* LAM *deciding $L(T)$ with* $\text{ram}(O(s))$ *memory, $C = O(1)$ blocks and a single deterministic oracle, with $N = O(t)$ oracle calls, $Q_{\text{blk}} = O(t)$ block movements, $Q_{\text{move}} = 0$, and $T_{\text{in}} = O(t \cdot b)$.*

*Proof.* Take $T$ with a one-way infinite tape, store it in ram from address 1, one symbol block per cell, keep the head position in the address register, and encode the finite TM state in controller control. Cells beyond the input are never initialised: a READ of an unwritten cell delivers the blank constant (Definition 3), which stands for $T$'s blank symbol, and the head never moves left of address 1, so the clipping of SHIFT is never exercised. The oracle receives the current tape symbol with an opcode for state $q$, returns the new symbol block $\langle \sigma' \rangle$, and emits a tag $\gamma = \langle q', L/R/\text{ halt} \rangle$. One step is: Read at the head address (one block movement), one oracle call, write the returned symbol block (one block movement), update finite control to $q'$, and SHIFT($\pm 1$) the address as selected by $\gamma$. At most the input symbol and returned symbol co-reside, so $C = 2$ suffices. SHIFT is a convenience, not a necessity: without it the controller keeps the head position as a block in a third slot, obtains its successor or predecessor by a second oracle call and applies SetAddr to the result, still $O(1)$ calls and movements per step, at $C = 3$. Everything the controller does is copying, addressing and finite branching, so vpa holds. Summing over $t$ steps gives the claim. Universality per se is not new [8]; the content here is that it costs $\Theta(1)$ calls and $\Theta(1)$ block movements per simulated step at constant context. □

## B. Context–memory communication

### B.1. Proof of Theorem 1

We use the red–blue convention of Section 4. Red pebbles model the active semantic store of capacity $C$, and blue pebbles slow memory of capacity $S$ ($S \geq n^2$ suffices for the grid of Corollary 11, whose lower bounds hold for every smaller $S$ as well). Initially every source, constants included, carries a blue pebble and nothing carries a red one. The legal moves are Compute (place a red pebble on a node all of whose predecessors carry red pebbles), Load (place a red pebble on a node carrying a blue pebble), Store (place a blue pebble on a node carrying a red pebble), and Delete (remove either pebble). At most $C$ red pebbles exist at any time, so a Compute output is inserted while its predecessors remain red; recomputation is allowed. The I/O cost is the number of Load and Store moves and $\mathrm{IO}_{C,S}(G)$ is its minimum over pebblings that end with every designated sink red or blue ($\mathrm{IO}_C$ when $S$ is not binding); a terminal red pebble on a sink is not charged a store. Other pebbling conventions require a corresponding change to the machine semantics and are not claimed here.

Fix a memory-initialized vertex-granular $G$-evaluation. Its routing, copying, memory management, and recomputation are unrestricted; the only semantic restriction is that each value-producing call realizes one declared vertex operation. The proof quotients physical block occurrences by their vertex provenance. No Const and no Input: every block in $X$ was read from $M$ or returned by a call, and every block in $M$ was there initially or was written from $X$. Provenance identity: blocks with different vertex labels are different pebbles even when their contents agree, recomputations of the same vertex share one pebble label, and a source-valued block may not stand in for an internal vertex. Terminal delivery: the run ends with a block standing for each designated sink in $X$ or in $M$, and Emit is not used inside the run. If auxiliary semantic calls are present, include them as explicit vertices of an augmented provenance DAG $G^+$ and apply the same argument to $G^+$.

*Proof of (i), soundness.* Fix a halting vertex-granular $G$-evaluation $R$. By vpa, every semantic block is an input, a constant, or an $L_n$-call output corresponding to a vertex of $G$; its vertex label (Definition 3) is that vertex. The map from occurrence labels to vertex labels is the quotient this proof works with: recomputed blocks are distinct occurrences with the same vertex label, and copying preserves both. Define, using vertex labels, for each time $t$,

$$\mathrm{Red}(t) = \{v : \text{a block labelled } v \text{ occupies any active semantic slot at } t\},$$
$$\mathrm{Blue}(t) = \{v : \text{a block labelled } v \text{ occupies a persistent-memory position at } t\}.$$

The active semantic slots include named controller block-register slots, so $|\,\mathrm{Red}(t)\,| \leq C$ and there is no additional uncharged semantic fast storage. Now map events: (i) an $L_n$ call producing $v$ has the blocks labelled by $\mathrm{preds}(v)$ as its arguments, so $\mathrm{preds}(v) \subseteq \mathrm{Red}$ at that instant. The returned block is appended before eviction and the call is a legal Compute of $v$; $C \geq d + 1$ allows the $d$ predecessors and result to co-reside; (ii) a $\mathrm{Read}(a)$ moving a block labelled $v$ from RAM into an active semantic slot is a Load (legal: $v \in \mathrm{Blue}$, as the cell was written earlier or holds an input); (iii) a $\mathrm{Write}(a, \cdot)$ of a resident block labelled $v$ is a Store; (iv) an eviction of the last resident copy of a block labelled $v$, or the overwriting of its last stored copy, is a Delete of the corresponding pebble, and any other eviction or overwrite changes no pebble. A Load of a vertex already red and a Store of a vertex already blue are legal moves and are still counted. Block movements in (ii)–(iii) are precisely the events counted by $Q_{\mathrm{blk}}$; copying or renaming among active semantic slots does not create extra pebbles, and because the run uses no Const or Input, (i)–(iii) are the only ways a pebble appears. The mapping is stated for ram; for stream, persistent positions are tape cells, Move changes no pebble, and Read/Write at the head map to Load/Store as in (ii)–(iii). For stack, Push moves rather than copies: it is a Store followed by a Delete of the red pebble when the moved block was the last active copy of its vertex, and Pop is a Load followed by a Delete of the blue pebble when it was the last stored copy; when a write overwrites the last stored copy of another vertex, that blue pebble is deleted before the store. Each ordering respects both capacities and charges one I/O move per transfer, which is the sense in which (i) holds for every interface. The result is a legal pebbling from the initial all-blue-sources position with red bound $C$, exactly $N$ Compute moves and at most $Q_{\mathrm{blk}}$ I/O moves. By terminal delivery each designated sink is resident or stored at the end, so it ends red or blue. □

*Proof of (ii), completeness.* Let $\Pi = (\mu_1, \ldots, \mu_L)$ be a legal pebbling of the fixed $L$-realisable functional DAG $G$ with red bound $C$, $P$ Compute and $I$ I/O moves. Assign each blue-pebble lifetime interval of $\Pi$ (a node $v$ from a Store, or from time zero for a source, until that blue pebble is deleted) a ram address $a_v$ by a fixed static allocation: $\Pi$ is fixed and carries at most $S$ blue pebbles at any time, so greedy colouring of these intervals in order of their left endpoints uses at most $S$ distinct addresses; a deleted blue pebble's block stays in its cell, ignored, until the address is reused, and since at most $S$ addresses are ever used, $M$ never holds more than $S$ blocks (one address per node suffices when $S \geq |V|$). Build a controller whose control states are $q_0, \ldots, q_L$, one per move — a finite set, since $G$ and $\Pi$ are fixed — with $q_{\ell-1} \to q_\ell$ executing $\mu_\ell$: Compute$(v) \mapsto$ the fixed oracle call realising $\phi_v$ on the resident predecessor blocks with its allowed opcode; Load$(v) \mapsto$ Read$(a_v)$; Store $(v) \mapsto$ Write$(a_v, \cdot)$; Delete$(v) \mapsto$ evict. Before this, drop from $\Pi$ every Load or Compute of a vertex already red: under set semantics such a move changes no pebble, so legality is preserved and neither $P$ nor $I$ increases, and every remaining Load or Compute finds a free slot, since a legal placement on a vertex not yet red requires fewer than $C$ red pebbles at that instant. Legality of $\Pi$ then guarantees each precondition holds and that at most $C$ blocks are ever resident. The controller selects blocks by fixed semantic slots and fixed addresses, so it is value-oblivious and satisfies vpa. The insertion convention matches Compute, including recomputation. The run is a memory-initialized vertex-granular $G$-evaluation: the sources sit at their addresses from the start, the context starts empty, no Const or Input is used, and since $\Pi$ ends with every sink red or blue the run ends with a block for each sink resident or stored. It has $N \leq P$ and $Q_{\text{blk}} \leq I$. The constructed machine is a ram machine, which is why the minimum is attained by one; combining with (i), which maps any interface's memory operations to loads and stores, gives $\min Q_{\text{blk}} = \text{IO}_{C,S}(G)$ over memory-initialized vertex-granular $G$-evaluations of this fixed realisation. □

**Corollary 9** (Priced call–traffic optimum). *For every $\lambda_N, \lambda_Q \geq 0$, $\min_R\{\lambda_N N_R + \lambda_Q Q_{\text{blk},R}\} = \min_\Pi\{\lambda_N P_\Pi + \lambda_Q I_\Pi\}$, where $R$ ranges over memory-initialized vertex-granular $G$-evaluations of $G$ with capacities $(C, S)$ and $\Pi$ over legal red–blue pebblings of $G$ with at most $C$ red and $S$ blue pebbles, $P_\Pi$ being the number of Compute moves and $I_\Pi$ the number of Load and Store moves; a ram memory-initialized vertex-granular $G$-evaluation attains the optimum, and one side is feasible if and only if the other is (an infeasible minimum being $+\infty$).*

*Proof.* For any memory-initialized vertex-granular $G$-evaluation $R$, the proof of (i) produces a legal pebbling with exactly $P_\Pi = N_R$ Compute moves and $I_\Pi \leq Q_{\text{blk},R}$ I/O moves, so the minimum pebbling objective is at most the minimum machine objective. Conversely, the proof of (ii) turns any legal $\Pi$ into a ram memory-initialized vertex-granular $G$-evaluation (one machine per pebbling, non-uniform as in the theorem) with $N_R \leq P_\Pi$ and $Q_{\text{blk},R} \leq I_\Pi$, so the minimum machine objective is at most the minimum pebbling objective. Nonnegative weights preserve both inequalities, giving equality, attained by the constructed ram machine; feasibility is equivalent by the same two directions. □

**Corollary 10** (Simultaneous call and transfer budgets). *Under the hypotheses of Theorem 1, for every $B_N, B_Q \geq 0$ a memory-initialized vertex-granular $G$-evaluation of $G$ with capacities $(C, S)$, $N \leq B_N$ and $Q_{\text{blk}} \leq B_Q$ exists if and only if a legal red–blue pebbling with at most $C$ red and $S$ blue pebbles, $P_\Pi \leq B_N$ Compute moves and $I_\Pi \leq B_Q$ I/O moves exists; a ram evaluation suffices whenever either side is feasible. Hence the sets of feasible budget pairs, and therefore their nondominated (Pareto-minimal) integer resource vectors $(N, Q_{\text{blk}})$ and $(P_\Pi, I_\Pi)$, coincide.*

*Proof.* The proof of (i) maps an evaluation with $(N, Q_{\text{blk}})$ to a pebbling with $P_\Pi = N \leq B_N$ and $I_\Pi \leq Q_{\text{blk}} \leq B_Q$; the proof of (ii) maps a pebbling with $(P_\Pi, I_\Pi)$ to a ram evaluation with $N \leq P_\Pi \leq B_N$ and $Q_{\text{blk}} \leq I_\Pi \leq B_Q$. The two feasible-budget sets are upward closed and coincide, so their minimal elements coincide; a weighted objective exposes only the Pareto points on the lower convex hull, whereas budgets expose every minimal element. □

**Remark 1** (Why the class is needed). Both clauses that the class adds to "evaluated vertex by vertex" are load-bearing. If Const were allowed, a constant source could enter the context at no charge while the game requires a Load for it, so (i) would fail by up to $|K_0|$ per use. If outputs could be streamed, a run could emit a sink and evict it: with one source, $C = 2$, $S = 1$ and four unary sinks, a run loads the source once and computes, emits and evicts each sink, for $Q_{\text{blk}} = 1$, whereas no pebbling with three pebble positions ends with four sinks red or blue. Such runs are matched by extending the game with an Output move that records a red-pebbled sink and may then delete its pebble; (i) and (ii) then hold verbatim for runs that Emit each sink once, with $\text{IO}_C$ replaced by the minimum over pebblings that output every sink. The grid bound of Corollary 11 holds for that game too, since its proof uses only that every internal vertex receives a Compute and that every source with a successor is loaded; we do not claim the matrix bound for it. Input-delivered vertex-granular $G$-evaluations (Definition 2), over which Theorem 3 states its $S = 0$ bound, are covered by (i) with each Input mapped to the placement of a pebble on the delivered source: in the black pebble game this is an ordinary placement (Appendix C), and in the red–blue game, with a blue bound of at least the number of sources, it is a Load of that source, so (i) holds with at most $Q_{\text{blk}}$ plus the number of sources I/O moves. Only direction (i) is claimed for this class: a legal pebbling may load a source more than once or, with $S = 0$, use no blue pebble at all, and neither is an input-delivered run. Theorem 3 needs no converse, its $S = 0$ upper bound being an explicit input-delivered schedule.

**Remark 2** (Necessity of value provenance). If the controller may compute an arbitrary polynomial-time function of the blocks it reads, it can emit the answer itself after reading the input once: for the ordinary matrix-multiplication DAG on $n \times n$ inputs this costs $N = 0$ model calls and $Q_{\text{blk}} = \Theta(n^2)$ transfers, against $\text{IO}_C(G) = \Omega(n^3/\sqrt{C})$. Input reads stay charged; what collapses is the value creation. The $\geq$ direction of Theorem 1, and with it results that reduce to pebbling, can collapse.

VPA forbids uncharged value creation; if parsing, retrieval scoring or code execution creates values, it must be modelled and charged separately.

## B.2. Corollary 11 and Proposition 12: statements and proofs

We give the full statements and proofs of the communication consequences used in Section 4.

**Corollary 11** (Communication bounds). *Let the $n \times n$ grid DAG have vertices $(i, j)$, $0 \le i, j < n$, sources on row 0 and column 0, sink $(n-1, n-1)$, and edges into each internal vertex $(i, j)$, $i, j \ge 1$, from $(i-1, j)$ and $(i, j-1)$. Over memory-initialized vertex-granular G-evaluations with $C \ge 3$ and $S \ge n^2$, the minimum transfer count is $Q_{\mathrm{blk}} = \Theta(n^2/C + n)$, hence $\Theta(n^2/C)$ for $C \le n$. For the ordinary $n^3$-operation matrix-multiplication DAG on $n \times n$ inputs, with $S \ge 2n^2$, enough to hold all sources initially (a memory-initialized vertex-granular G-evaluation also needs $C + S \ge 2n^2 + n$; Appendix B), $Q_{\mathrm{blk}} = \Omega(n^3/\sqrt{C})$.*

Both orders are bounds of J.-W. Hong and H. T. Kung [18]; the grid is reproved below with explicit constants, recomputation allowed, and matched by tiles of side $\Theta(C)$. Both bound evaluation of the prescribed DAG, not computation of its function by other means.

**Proposition 12** (Sequential input traffic). *Consider an $n$-step chain in which each step produces one semantic block of width $b$. If the $i$-th call receives all earlier blocks, its input size is $\Theta(ib)$ and $T_{\mathrm{in}} = \Theta(n^2 b)$. If each step instead depends on only $O(1)$ earlier blocks and only those blocks are supplied, then $T_{\mathrm{in}} = \Theta(nb)$, and no LAM evaluating a non-degenerate chain vertex by vertex, each call taking the previous block as an argument, has $T_{\mathrm{in}} < nb$.*

*Proof of Corollary 11.* Both DAGs have in-degree at most two, so the insertion convention needs only $C \ge 3$, and Theorem 1, whose hypotheses on memory-initialized vertex-granular $G$-evaluations the corollary inherits, reduces the claim to bounds on $\mathrm{IO}_C$ in the red–blue game of Appendix B, recomputation allowed, with $S$ at least the number of sources, as the corollary assumes; the lower bounds hold a fortiori for smaller $S$, and the schedules of the upper bounds hold at most $(2n-1) + (n-1)^2 = n^2$ blue pebbles. Sources carry blue pebbles initially, and a source can only become red by a LOAD.

*Grid, lower bound.* Fix a legal pebbling with $q$ I/O moves. Cut its move sequence after every $C$-th I/O move; this gives $h = \lfloor q/C \rfloor + 1 \le q/C + 1$ consecutive phases, each containing at most $C$ I/O moves. Let $V_i$ be the set of internal vertices that receive a COMPUTE in phase $i$, and let $F_i$ be the set of vertices outside $V_i$ with an edge into $V_i$.

*Claim 1:* $|F_i| \le 2C$. Let $u \in F_i$ with an edge $u \to v$, $v \in V_i$. At the COMPUTE of $v$ in phase $i$, $u$ is red. A red pebble on $u$ is placed by COMPUTE($u$), which does not occur in the phase since $u \notin V_i$, or by LOAD($u$); so either $u$ was red at the start of the phase, or it was loaded during the phase. At most $C$ vertices are red at the start and at most $C$ loads occur in the phase.

*Claim 2:* $|V_i| \le |F_i|^2$. Let $R$ be the set of rows and $J$ the set of columns that meet $V_i$. In a row $r \in R$ let $(r, j_r)$ be the leftmost vertex of $V_i$; it is internal, so $j_r \ge 1$ and $(r, j_r - 1)$ is an in-neighbour of it outside $V_i$, hence in $F_i$. Distinct rows give distinct such vertices, so $|R| \le |F_i|$; the topmost vertex of each column gives $|J| \le |F_i|$; and $|V_i| \le |R||J|$.

Every internal vertex is computed in some phase, since the sink must receive a pebble and a non-source vertex receives its first pebble only by COMPUTE, which needs its predecessors red. Hence $(n-1)^2 \le \sum_i |V_i| \le 4hC^2$, so $h \ge (n-1)^2/(4C^2)$ and $q \ge C(h-1) \ge (n-1)^2/(4C) - C$. For $C \le (n-1)/4$ the subtracted term is at most $(n-1)^2/(16C)$, giving $q \ge \frac{3}{16}(n-1)^2/C$. Independently, each of the $2n-2$ sources is a predecessor of an internal vertex and must be loaded once, so $q \ge 2n-2$. Together, for every $C \ge 3$ and $n \ge 2$,

$$\mathrm{IO}_C \ge \max\left\{ \frac{3}{16}\frac{(n-1)^2}{C}\,\mathbf{1}[C \le \frac{n-1}{4}], 2n-2 \right\} \ge \frac{3}{128}\left(\frac{n^2}{C} + n\right),$$

where the last inequality uses $n - 1 \ge n/2$ and, for $C > (n-1)/4$, $n^2/C \le 8n$. This is the Hong–Kung partition argument with the immediate-predecessor frontier $F_i$ in place of a dominator set; Claim 2 is the isoperimetric step, and neither claim uses that a vertex is computed only once.

*Grid, upper bound.* For $4 \le C \le n$ put $s = \lfloor (C-2)/2 \rfloor \ge 1$ and cover the internal vertices by $s \times s$ tiles (smaller at the boundary) taken in row-major order of tiles. For a tile, load its north halo (the row above it, $\le s$ values) and west halo (the column to its left, $\le s$ values): these are sources or values stored by the tile above or to the left. Compute the tile row by row, left to right, discarding a halo value or a computed value once its last use in the tile has occurred, and store the tile's south row and east column ($\le 2s$ values). While computing $(r, c)$ the resident blocks are the unused north-halo or previous-row values in columns $\ge c$, the current-row values in columns $< c$, the unused west-halo values, and the output: at most $s + s + 1 = 2s + 1 \le C - 1$. Each tile costs at most $4s$ transfers, and there are at most $\lceil (n-1)/s \rceil^2$ tiles, so

$$\mathrm{IO}_C \le 4s\left(\frac{n-1}{s} + 1\right)^2 \le \frac{4(n-1)^2}{s} + 8n + 4s \le \frac{32n^2}{C} + 10n,$$

using $s \geq (C-3)/2 \geq C/8$ for $C \geq 4$ and $4s \leq 2(C-2) \leq 2n$. At $C = 3$ the row-major schedule, which loads two predecessors and stores one result per vertex, gives $\mathrm{IO}_3 \leq 3(n-1)^2$. For $C \geq n+2$ the row-major schedule keeps the previous row resident, loads each of the $2n-2$ sources with a successor once and leaves the sink resident, so $\mathrm{IO}_C \leq 2n-2$, meeting the source count of the lower bound; for $n < C < n+2$ use $\mathrm{IO}_C \leq \mathrm{IO}_n$. Hence $\mathrm{IO}_C = O(n^2/C + n)$ for every $C \geq 3$, and with the lower bound $\mathrm{IO}_C = \Theta(n^2/C + n)$. The tiled schedule is uniform in the controller model of Definition 3: row, column and tile indices, and hence all addresses, are public structural counters, so this is a family-level construction rather than an appeal to the non-uniform completeness direction of Theorem 1.

*Matrix multiplication.* The ordinary DAG has one multiplication vertex per triple $(i, j, k)$ and adds the $n$ products of each entry along a chain, so its in-degree is two; J.-W. Hong and H. T. Kung [18] prove $\mathrm{IO}_C \geq a\,n^3/\sqrt{C}$ for it, for an absolute constant $a > 0$ and all $C \leq c_0 n^2$ with a constant $c_0 > 0$ (see also G. Ballard, J. Demmel, O. Holtz, and O. Schwartz [38]); we carry $a$ symbolically rather than fix its value. If the game in which the bound is proved requires every output to end blue rather than red or blue, at most $C$ additional stores turn a pebbling of our game into one of that game, so the bound transfers with an additive loss of at most $C$. Put $\eta = \min\left\{c_0, (a/2)^{2/3}\right\}$. For $C \leq \eta n^2$ one has $C^{3/2} \leq (a/2)n^3$, i.e. $C \leq (a/2)n^3/\sqrt{C}$, so $\mathrm{IO}_C \geq (a/2)n^3/\sqrt{C}$; for $C > \eta n^2$ the $2n^2$ input loads alone give $\mathrm{IO}_C \geq 2n^2 > 2\sqrt{\eta}\,n^3/\sqrt{C}$. Both cases are $\Omega(n^3/\sqrt{C})$ with a constant depending on $a$ and $c_0$ only. This is a bound for evaluating that DAG, not for computing the product by another algorithm. The hypothesis $S \geq 2n^2$ is necessary for initial source storage, but does not by itself guarantee feasibility at every context capacity; the lower bound applies whenever a memory-initialized vertex-granular $G$-evaluation exists. (A memory-initialized vertex-granular $G$-evaluation of this DAG needs $C + S \geq 2n^2 + n$: a source cannot be regenerated, so just before the first deletion of a source's last copy all $2n^2$ sources still hold slots, and each of the $n$ vertex-disjoint paths from that source to $n$ distinct outputs must already hold a non-source block, $2n^2 + n$ blocks in $C + S$ slots; if no source is ever deleted, the $n^2$ output blocks of the final configuration join the $2n^2$ sources instead. At $n = 4$, $C = 3$, $S = 32$ this is 36 blocks in 35 slots, so no memory-initialized vertex-granular $G$-evaluation exists although $S = 2n^2$.) We make no transfer claim here for adaptive comparison sorting: its external-memory lower bound uses a different model and parameter mapping. □

*Proof of Proposition 12.* Index the chain by an initial block $x_0$, held in $M$ from the start (if it is instead delivered once by Input, one further Write puts it there), and calls $1, \ldots, n$, call $i$ producing $x_i$. *Replay.* Call $i$ takes the $i$ blocks $x_0, \ldots, x_{i-1}$ as arguments, so with the inserted output it needs $i+1$ slots, $C \geq n+1$ at call $n$, and $T_{\mathrm{in}} = \sum_{i=1}^{n} i \cdot b = n(n+1)b/2 = \Theta(n^2 b)$. *Persistent memory.* The controller loads $x_{i-1}$ (one block movement), calls the oracle with the two blocks $(x_{i-1}, \langle\text{step}\rangle)$, three slots with the inserted output, and stores $x_i$ (one block movement): $C = 3$, $T_{\mathrm{in}} = 2nb$, $Q_{\mathrm{blk}} = 2n$ and $Q_{\mathrm{move}} = 0$. *Lower bound.* Assume each chain step is non-degenerate, i.e. $x_i$ is not a copy of any earlier block. Then every $x_i$, $i \geq 1$, is the output of a distinct model call: under vertex-wise evaluation it is produced by a call, and vpa rules out its manufacture by the controller; so $N \geq n$, and since that call takes the block standing for $x_{i-1}$ as an argument, it carries at least $b$ symbols and $T_{\mathrm{in}} \geq nb$. Definition 3 permits zero-argument calls, which is why the proposition names this hypothesis. Hence replay is a factor $\Theta(n)$ from optimal in prompt tokens. □

# C. Memory architecture

## C.1. Proof of Theorem 2

*Upper bound.* The ram controller keeps the current index in the address register and repeats $k$ times: Read$(a)$ into its one slot, then SetAddr on that slot, which decodes the stored index $\pi(a)$ into the address register, then evict; after the $k$-th read it emits the block instead. Per hop there is one block read and no head motion, so $N = 0$, $Q_{\mathrm{blk}} = Q = k$, $Q_{\mathrm{move}} = 0$, $Q_{\mathrm{obs}} = k$ and $C = 1$. If decoding the entry is instead charged as an oracle call, its returned block is the argument of SetAddr, giving $N = k$ and $C = 2$; a tag cannot carry the index in a uniform family, $\Gamma$ being fixed as $n$ grows, whereas the instancewise lower bounds place no bound on $\Gamma$ and use none.

*Lower bound.* Work in the read-only stream model of the theorem: cell $i$ contains $\pi(i)$, the head starts at cell $a_0 = 1$, and

$$Q = Q_{\mathrm{mem}} = Q_{\mathrm{blk}} + Q_{\mathrm{move}}$$

(the appendices write $Q$ for the main text's $Q_{\mathrm{mem}}$), where reads contribute to $Q_{\mathrm{blk}}$ and unit head displacement contributes to $Q_{\mathrm{move}}$. Put $a_{j+1} = \pi(a_j)$ and let $D_j$ be the event that $a_0, \ldots, a_j$ are distinct.

Non-speculation fixes the order of first reads: a cell may be read only once its address has been revealed, and after the first read of cell $a_{i-1}$ the revealed addresses are $a_0, \ldots, a_i$, so the first reads of new cells occur in the order $a_0, a_1, a_2, \ldots$ Correctness on every permutation forces these reads to happen. Fix $\pi \in D_j$ with $1 \leq j \leq k-1$ and suppose the run on $\pi$ never reads cell $a_j$; then it reads only cells among $a_0, \ldots, a_{j-1}$. The machine is deterministic and, by the interface clause of Definition 3, every call returns a fixed function of its arguments, so the run, and with it the output, is the same on every permutation $\pi'$ with $\pi'(a_i) = a_{i+1}$ for $i < j$, whatever its constants or tags. Among these permutations one has $\pi'(a_j) = a_0$, so that the orbit of $a_0$ is the cycle $a_0, \ldots, a_j$ and the correct answer $a'_k$ lies in $\{a_0, \ldots, a_j\}$; another continues the orbit $a_0, \ldots, a_j$ through $k -$ $j$ further distinct cells, which exist since $k+1 \le n$, so that $a'_k$ lies outside $\{a_0, \dots, a_j\}$. The machine answers the same on both and errs on one, contradicting always-correctness. Hence on $D_j$ the run first reads $a_0, a_1, \dots, a_j$ in this order, the first read of $a_i$ after that of $a_{i-1}$; the head may pass over a cell without reading it, but it must travel from each first read to the next, one MOVE per cell, since Definition 3 gives a STREAM machine no SETADDR; so with $J$ the largest index $j \le k-1$ for which $D_j$ holds,

$$Q_{\text{move}} \ge \sum_{j=1}^{J} |\, a_j - a_{j-1} \,| = \sum_{j=1}^{k-1} |\, a_j - a_{j-1} \,| \, \mathbf{1}_{D_j}.$$

For a uniformly random permutation, conditional on $D_{j-1}$ and on the values $a_0, \dots, a_{j-1}$, the next value $a_j = \pi(a_{j-1})$ is uniform on the $n-j+1$ indices outside $\{a_1, \dots, a_{j-1}\}$. At most $n/4+1$ indices lie at distance less than $n/8$ from any given point, so for $j \le k \le n/4$ at least $(n-j+1)-1-(n/4+1) \ge n/2-1 \ge n/4$ of those indices differ from $a_0$ and lie at distance at least $n/8$ from $a_{j-1}$; therefore $\mathbb{E}[|\, a_j - a_{j-1} \,| \, \mathbf{1}_{D_j} \mid D_{j-1}, a_0, \dots, a_{j-1}] \ge \frac{n/4}{n-j+1} \cdot \frac{n}{8} \ge \frac{n}{32}$. The cycle of $a_0$ has length uniform on $[n]$, so $\Pr[D_{j-1}] \ge \Pr[D_k] = 1 - k/n \ge 3/4$. Summing over $j \le k-1$, $\mathbb{E}_\pi[Q_{\text{move}}] \ge \frac{3}{4} \cdot \frac{n}{32}(k-1) \ge \frac{3}{256}kn$ for $k \ge 2$, and $Q \ge Q_{\text{move}}$. Consequently some permutation incurs this aggregate memory-interface-step cost. It is dominated by sequential head motion and is not an $\Omega(knb)$ lower bound on physically transferred symbols; only $Q_{\text{blk}}$ is semantic block traffic. A full-table scan made before an index is requested is excluded by non-speculation; the proof does not charge such a scan once per future hop.

*Randomised machines.* Let the machine have a random tape and be correct on every permutation for almost every tape. For every tape outside a null set (the intersection, over the finitely many permutations, of probability-one events) the machine with that tape fixed is a deterministic, always-correct, non-speculative machine of the kind just treated: the interface clause is unchanged, and "the run is the same on $\pi$ and $\pi'$" holds tape by tape. Hence $\mathbb{E}_\pi[Q_{\text{move}}] \ge \frac{3}{256}kn$ for almost every tape, and averaging over the tape gives the bound of the theorem; the distance sum itself does not depend on the tape, only the event that the run reads $a_0, \dots, a_J$ in order does.

**Remark 3** (Speculative access: workspace, shortcuts and a protocol reduction)**.** The theorem is for the non-speculative class. For speculative machines the answer depends on every store Definition 3 leaves unbounded; we list the escapes, then the class in which none is available, and then an algorithm inside that class showing that $\Omega(kn)$ is the wrong question even there. Bounding the context alone does not suffice: for one fixed $n$, Definition 3 allows $|\,\Gamma\,| \ge n$ and $|\,Q_{\text{ctrl}}\,| \ge n!$, and then a $C=1$ machine scans the tape once ($n-1$ MOVES, $n$ READS), passes each cell's block to a verifier whose tag is the index it holds ($Q_{\text{obs}} = n$), records $\pi$ in its finite control, computes $a_{k-1}$ there, moves the head to cell $a_{k-1}$ (at most $n-1$ MOVES), reads it and emits the block: $Q_{\text{move}} \le 2n-2$ for every $k$. Bounding the control alone does not suffice either: with $C \ge n$ and $|\,\Gamma\,| = n$, one scan reads every cell into the context, and each hop then selects the next slot from the tag returned by a verifier call on the current block, with no further head motion. Bounding the block width is necessary as well: with $b(n) \ge n\lceil \log_2 n \rceil$, a $C=3$ machine with $O(1)$ control scans the tape once, folding each cell into one accumulator block by a model call, and obtains $a_k$ from the accumulator by one more call, so $Q_{\text{move}} = n-1$ for every $k$. Finally, the content-dependent state of Definition 3 is not the registers alone: the step counter, which $\tau_{\text{ctrl}}$ reads and can advance by waiting, holds $\log_2 t$ bits, unbounded when the running time is. A machine with $C=1$, $|\,\Gamma\,| = n$ and $O(n)$ control states writes $\pi$ into the counter during one scan, one base-$(n+3)$ digit per cell, by waiting after reading cell $i$ until digit $n-i+1$ of the counter equals the tag just read (that digit is 0 when the wait begins and the steps taken since the previous wait are fewer than $n+3$, so the wait ends exactly when the counter reads the digits written so far followed by zeros; the lowest digit absorbs the final walk); it then computes $a_{k-1}$ from the counter, walks to it and reads: $Q_{\text{move}} \le 2n-2$ in $n^{O(n)}$ steps. Fifth, a writable tape: with $\Theta(n)$ blank writable cells beside the table, the arrangement of blocks on them is $\Theta(n \log n)$ bits of state that no count of registers, pattern and counters sees, and pointer doubling by external-memory sorting (a merge pass costs $O(n^2/C)$ head moves with $C$ blocks of context) reaches $a_k$ in $O((n^2/C)\log(n/C)\log k)$ moves, $o(kn)$ at $C = n/\log n$ once $k \ge \log^2 n$.

The *bounded-workspace class* closes all five. It consists of the always-correct speculative STREAM LAMS of Definition 3 on the task as stated, one index per block with $b(n) = \lceil \log_{|\,\Sigma\,|} n \rceil$, whose tape holds the $n$ table cells and nothing else ($S = n$) with WRITE unavailable, with $C = o(n)$ context slots, $h = O(\log n)$ register bits (so $|\,Q_{\text{ctrl}}\,|\,|\,\Gamma\,| = n^{O(1)}$, which every uniform family on this task satisfies) and $n^{O(1)}$ steps, so that the counters hold $O(\log n)$ bits; tool and verifier calls are unrestricted in number, each a fixed map of its arguments. Such a machine holds $O(C \log n)$ bits of content-dependent state in all (context, registers, pattern and counters), fewer than the $\log_2 n!$ bits of $\pi$, so it cannot memorise the table in a scan.

$\Omega(kn)$ nevertheless fails in this class for large $k$, because $C$ blocks can hold *shortcuts* along $\pi$ rather than $\pi$ itself. Fix $2 \le m \le C/4$ landmark cells including cell 1 (so $C \ge 12$, which makes $3m+3 \le C$). For each landmark the machine keeps three blocks: its index, a walker position (initially the landmark) and a hop count, stored as the index block $\langle \text{count}+1 \rangle$, initially $\langle 1 \rangle$; the count never exceeds $n-1$, because with $m \ge 2$ landmarks the next landmark on an orbit is at most $n-1$ hops away, so it fits one index block. It also keeps one block $\langle c\rangle$ equal to the head position, updated by an increment or decrement tool call at each Move, one working slot, and one accumulator slot, which is free during the sweeps and holds the remaining hop budget in the query phase: $3m+3\le C$ slots in all. A tool call's output takes a free slot, the accumulator slot during the sweeps and the working slot in the query phase, and is then moved into place by Evict and Copy; index blocks are built from the constant $\langle 1\rangle$ by the same tool. The target $k$, a public parameter of the task, is held in the control's $O(\log n)$ bits and enters the blocks only through the accumulator, initialised to $\langle k+1\rangle$ by $k$ increment calls, the remaining budget being stored with the same offset of one as the counts. No status bit lives in the finite control: a walker is *finished* exactly when its position is a landmark and its count is positive, a fixed map of its three blocks and the $m$ index blocks that a verifier call reports as a tag, and one verifier call on all $3m$ blocks reports whether every walker is finished. The machine sweeps the tape back and forth, reading that tag once per sweep and stopping when it is set. At cell $c$ it reads the cell into the working slot and, for each walker in turn (the control holds the walker index, $O(\log n)$ bits), calls a verifier on $\langle c\rangle$ and the walker's three blocks; when the tag reports an unfinished walker at position $c$, the walker's position block is replaced by the block just read and its count block by the incremented count, the tool output landing in the accumulator slot before being copied into place; at most one unfinished walker matches a cell, the walkers' arcs on a cycle being disjoint. A sweep advances every unfinished walker at least once, so after $G$ sweeps, $G$ the largest number of hops from a landmark to the next landmark on its orbit, the context holds for every landmark the next landmark and the distance to it. The query follows these shortcuts from $a_0=1$, subtracting each shortcut's hop count from the accumulator by a tool call as long as the comparison verifier on the accumulator and the next count block reports that the shortcut does not overshoot the budget, then walks the remaining $r<G$ hops on the tape, each hop moving the head to the cell named by the current block: a comparison verifier on $\langle c\rangle$ and that block returns a tag in $\{<,=,>\}$, which gives the direction and says when to stop, so a hop costs at most $n$ Moves (an equality-only verifier would cost $2n$, a sweep to one end and back, changing no order below). Every call is a fixed map of its arguments, the machine is deterministic, uses no random tape and is always correct, and it uses $3m+3\le C$ slots, $O(\log n)$ register bits (the control indexes walkers, slots and opcodes) and $O(n^2m)$ steps. For uniformly random $\pi$ the orbit from a landmark visits fresh cells uniformly, so $\Pr[G>s]\le m(1-m/n)^s$ and $\mathbb{E}[G]=O((n/m)\log m)$, whence, for every $k$,

$$\mathbb{E}_\pi[Q_{\mathrm{move}}]\le 2n\,(\mathbb{E}[G]+1)=O\!\left(\frac{n^2}{C}\log C\right)$$

(at most $G$ sweeps of $n-1$ moves and $r<G$ hops of at most $n$ moves), which is $o(kn)$ whenever $k=\omega((n/C)\log C)$; at $C=\sqrt{n}$ and $k=n/4$ it is $O(n^{3/2}\log n)$ against $\Theta(n^2)$ for the non-speculative class. Together with the demand-following scan, which costs $O(kn)$ moves inside the class, the best bound exhibited is $\mathbb{E}_\pi[Q_{\mathrm{move}}]=O(\min\{kn,(n^2/C)\log C\})$ for $C\ge 12$.

The question for the bounded-workspace class is therefore the trade-off between context and head motion, not the exponent of $kn$. We conjecture that every *deterministic* machine of the class has $\mathbb{E}_\pi[Q_{\mathrm{move}}]=\Omega(\min\{kn,n^2/C\})$ up to a factor polylogarithmic in $n$: a scan costs $n$ moves and advances at most $O(C)$ walkers, and only walkers on the orbit of $a_0$ can help. Nothing is lost by stating the conjecture over deterministic machines: it concerns a fixed input distribution, and a randomised machine's expectation over $\pi$ and its tape is an average of deterministic machines' expectations over $\pi$. What is proved is the separation for the non-speculative class; what is exhibited is the speculative upper bound $O((n^2/C)\log C)$ in expectation over $\pi$; the remaining step is the converse, which the proof of the theorem cannot supply because a scan may read $\pi(j)$ before $j$ is revealed and retain it, and the sum of distances between successively revealed cells has no analogue once a $C$-block cache biases the next orbit location. Independently of the workspace bound, every always-correct machine, speculative or not, deterministic or zero-error randomised, has $\mathbb{E}[Q_{\mathrm{move}}]\ge n/9$ for $n\ge 16$ and $2\le k\le n/4$ (Proposition 14, Corollary 15).

The route to the converse is communication complexity, and the step from the machine to a protocol is in place (Lemma 13): a machine of the class holds $s=O(C\log n)$ bits of content-dependent state, and cutting the tape between cells $c$ and $c+1$ turns its run into a deterministic two-party protocol that sends that state once per crossing of the cut; since every Move that changes the head position crosses exactly one of the $n-1$ cuts, some cut in the middle half is crossed $O(\mathbb{E}_\pi[Q_{\mathrm{move}}]/n)$ times in expectation over the *same* uniform $\pi$, and stopping the protocol after $r$ messages gives an $r$-round protocol correct with probability $1-O(\mathbb{E}_\pi[Q_{\mathrm{move}}]/(nr))$; crossings of one cut, not passes over the tape, are what two-party round elimination consumes. The landmark algorithm is such a protocol with $r=O((n/C)\log C)$ expected rounds of $s=O(C\log n)$ bits, so $rs=\tilde{\Theta}(n)$, and relaying the orbit is one with at most about $k/2$ rounds of $\log n$ bits. The conjecture would therefore follow from a distributional round lower bound for this instance: that no deterministic protocol with $r=o(k)$ rounds of $s$ bits and $rs=o(n/\mathrm{polylog}\,n)$ outputs $\pi^k(1)$ with probability $3/4$ over a uniform permutation split by cell position. (A worst-case version on rectangles of density $2^{-O(rs)}$ is false: such a rectangle can force the orbit to stay on one side for $\Theta(rs)$ steps.) For the classical alternating instance the corresponding statements are the round-elimination bounds of N. Nisan and A. Wigderson [39]; in our instance the side of the next pointer is the position of a uniformly random value, so the orbit changes side at input-dependent times and the speaker often owns the current cell, a structure for which the round-elimination step remains to be carried out.

**Lemma 13** (Head moves are cut crossings). *Let $\Lambda$ be a deterministic always-correct machine of the bounded-workspace class of Remark 3 (on every $\pi$ it emits exactly one block, the one encoding $a_k$) on a linear tape of cells $1, \dots, n$ with no* Move *off either end, with $\Sigma$ a fixed alphabet, and let $s$ be the length of a fixed-length binary encoding of its* content-dependent configuration*: control state, tag, step count, occupancy pattern and the contents of the $C$ context slots ($s = O(C \log n)$ in the class). Let $\bar{Q} \coloneqq \mathbb{E}_\pi[Q_{\mathrm{move}}]$ over uniform $\pi$. For a cut $c \in [n-1]$ let $X_c(\pi)$ be the number of* Moves *between cells $c$ and $c+1$ on input $\pi$, and let $P_c$ be the two-party protocol in which Alice holds cells $1, \dots, c$, Bob holds cells $c+1, \dots, n$, the party whose cells contain the head simulates $\Lambda$, sends the content-dependent configuration to the other party whenever a* Move *crosses the cut, and, at the first* Emit*, announces the emitted block and stops. Let $I \coloneqq \{c \in \mathbb{Z} : n/4 \le c \le 3n/4\}$ and $n \ge 16$.*

1. *$P_c$ is deterministic, Alice speaks first, and on every $\pi$ it sends at most $X_c(\pi)$ messages of $s$ bits and outputs $a_k$; moreover $\sum_{c=1}^{n-1} X_c(\pi) \le Q_{\mathrm{move}}(\pi)$.*
2. *$\bar{Q} \ge \sum_{c \in I} \mathbb{E}_\pi[X_c]$, so some $c^* \in I$ has $\mathbb{E}_\pi[X_{c^*}] \le 4\bar{Q}/n$; stopping $P_{c^*}$ when an $(r+1)$-st message would be sent gives a protocol with at most $r$ messages that outputs $a_k$ with probability at least $1 - 4\bar{Q}/(n(r+1))$ over $\pi$.*
3. *If $c \in I$, $\mathbb{E}_\pi[X_c] \le (r+1)/4$ and $rs \le n/17 - 3$, then there are a $c$-subset $S \subseteq [n]$ with $|S \cap \{c+1, \dots, n\}| \ge n/16$, a set $A$ of bijections $\{1, \dots, c\} \to S$ with $|A| \ge 2^{-rs-2} c!$ and a set $B$ of bijections $\{c+1 \dots n\} \to [n] \setminus S$ with $|B| \ge 2^{-rs-2}(n-c)!$ such that on every $\pi$ with $(\pi|_{[c]}\ \pi|_{[n] \setminus [c]}) \in A \times B$ the protocol $P_c$ sends at most $r$ messages and outputs $a_k$.*

*Proof.* The configuration is encoded in $\lceil \log_2 |Q_{\mathrm{ctrl}}| \rceil + \lceil \log_2 |\Gamma| \rceil + \lceil \log_2(T_{\max} + 1) \rceil + C + C\, b(n) \lceil \log_2 |\Sigma| \rceil$ bits, padded to this fixed length, where $T_{\max}$ is the largest running time over the finitely many inputs; in the class $|Q_{\mathrm{ctrl}}|\,|\Gamma| = n^{O(1)}$, $T_{\max} = n^{O(1)}$ and $b(n) \lceil \log_2 |\Sigma| \rceil = O(\log n)$, so $s = O(C \log n)$. The head position is not part of the configuration and the tape is the input; the public counters are fixed functions of the step count and of $n$.

(1) We show by induction on $t$ that the active party can execute step $t$ of $\Lambda$ exactly. It knows the configuration after step $t-1$ (initially Alice: the initial configuration is fixed and the head is at cell 1) and the head position, which lies in its own cells. By Definition 3 the next control state and operation are $\tau_{\mathrm{ctrl}}$ of the registers, the counters and the occupancy pattern, all determined by the configuration and the head position. A Call is a fixed map of its argument blocks (interface clause), which are context blocks, so the active party evaluates it; Copy, Evict and Const act on the context only; a Read returns the block in the head cell, which the active party holds; Write is unavailable, and the ram, stack and Input operations do not apply. A Move that keeps the head on the same side of the cut changes nothing else; a Move from $c$ to $c+1$ or from $c+1$ to $c$ is the only event that generates a message: the active party sends the configuration after the move, and the receiver, who knows from the direction of the message that the head is now at $c+1$ (Alice sent) or at $c$ (Bob sent), continues the simulation. At the first Emit the active party announces the emitted block and the protocol stops. Hence the simulation reproduces the run of $\Lambda$ on $\pi$ up to its first Emit, and the announced block encodes $a_k$ by always-correctness. Messages are sent only at crossings of the cut before that step, so there are at most $X_c(\pi)$ of them, each of exactly $s$ bits; the head starts in Alice's cells and moves by unit steps, so the first crossing is from $c$ to $c+1$ and Alice speaks first, after which the speakers alternate. Finally, a Move changes the head position by $\pm 1$ or, if clipped at an end of the tape, not at all, so it crosses at most one of the boundaries $\{c, c+1\}$, $c \in [n-1]$, exactly one when the head moves; summing over $c$ gives $\sum_{c=1}^{n-1} X_c(\pi) \le Q_{\mathrm{move}}(\pi)$.

(2) Taking expectations in (1) and dropping the non-negative terms with $c \notin I$ gives $\bar{Q} \ge \sum_{c \in I} \mathbb{E}_\pi[X_c]$ ($I \subseteq [n-1]$ for $n \ge 4$). The number of integers in $[n/4, 3n/4]$ is at least $n/2 - 1 \ge n/4$ for $n \ge 4$, so $\min_{c \in I} \mathbb{E}_\pi[X_c] \le \bar{Q}/|I| \le 4\bar{Q}/n$. The stopped protocol is deterministic, Alice first, sends at most $r$ messages, and coincides with $P_{c^*}$ whenever $X_{c^*}(\pi) \le r$, in which case it outputs $a_k$ by (1); as $X_{c^*}$ is a non-negative integer, Markov's inequality gives $\Pr_\pi[X_{c^*} \ge r+1] \le \mathbb{E}_\pi[X_{c^*}]/(r+1) \le 4\bar{Q}/(n(r+1))$.

(3) For a transcript $\tau$ let $R_\tau$ be the set of $\pi$ on which $P_c$ produces $\tau$. Alice's messages, and her decision to stop, are functions of $\pi|_{[c]}$ and of the previous messages, Bob's of $\pi|_{[n] \setminus [c]}$ and the previous messages, so $R_\tau = \left\{\pi : \pi|_{[c]} \in \mathscr{A}_\tau\ \pi|_{[n] \setminus [c]} \in \mathscr{B}_\tau\right\}$ for some sets $\mathscr{A}_\tau, \mathscr{B}_\tau$, and $P_c$ outputs $a_k$ on all of $R_\tau$ by (1). On $\{\pi : X_c(\pi) \le r\}$, an event of probability at least $3/4$ by Markov's inequality, $P_c$ sends at most $r$ messages, so the sets $R_\tau$ with at most $r$ messages cover it; there are at most $\sum_{i=0}^{r} 2^{is} \le 2^{rs+1}$ such transcripts, so some $R_\tau$ has probability at least $\frac{3}{8} 2^{-rs}$. Condition on $S \coloneqq \pi([c])$, a uniformly random $c$-subset of $[n]$: given $S$, the restrictions $\pi|_{[c]}$ and $\pi|_{[n] \setminus [c]}$ are independent uniform bijections onto $S$ and $[n] \setminus S$, and $R_\tau \cap \{\pi([c]) = S\} = A_S \times B_S$ with $A_S := \{p \in \mathscr{A}_\tau : p([c]) = S\}$ and $B_S := \{\bar{p} \in \mathscr{B}_\tau : \bar{p}([n] \setminus [c]) = [n] \setminus S\}$. The count $|\, S \cap \{c+1, \ldots, n\} \,|$ is hypergeometric with mean $c(n-c)/n \geq 3n/16$ for $c \in I$, so by Hoeffding's inequality for sampling without replacement [40] it is below $n/16 \leq 3n/16 - n/8$ with probability at most $\exp(-2(n/8)^2/c) \leq e^{-n/24} \leq 2^{-n/17}$, which is at most $\frac{1}{8}2^{-rs}$ when $rs \leq n/17 - 3$. Hence the subsets $S$ with $|\, S \cap \{c+1, \ldots, n\} \,| \geq n/16$ carry at least $(\frac{3}{8} - \frac{1}{8})2^{-rs}$ of the probability of $R_\tau$, and one of them has $\Pr[A_S \times B_S \mid S] = \frac{|\, A_S \,|}{c!} \cdot \frac{|\, B_S \,|}{(n-c)!} \geq \frac{1}{4}2^{-rs}$; both factors are at most 1, so each is at least $2^{-rs-2}$. Take $A := A_S$ and $B := B_S$. □

The density $2^{-rs-2}$ in (3) is too small for a worst-case round lower bound on such rectangles: a rectangle of density $2^{-O(k)}$ can force the orbit of 1 to stay on one side of the cut for $k$ steps, after which one message suffices.

**Proposition 14** (Every deterministic always-correct machine moves $\Omega(n)$)**.** *Let $\Lambda$ be a deterministic always-correct* stream lam *for the pointer-chasing task on a read-only linear tape with $S = n$ and the head initially at cell 1, with no restriction on $C$, the control, the tags, the counters, the block width or the calls, and let $n \geq 16$ and $2 \leq k \leq n/4$. Let $D_k$ be the event that $a_0, \ldots, a_k$ are distinct. Then $Q_{\text{move}}(\pi) \geq \min\{a_{k-1} - 1, n-k-3\}$ for every $\pi \in D_k$, and $\mathbb{E}_\pi[Q_{\text{move}}] \geq \frac{243}{2048} n \geq n/9$.*

*Proof.* Let $R(\pi) \subseteq [n]$ be the set of cells read during the run on $\pi$. *(i) The run depends on $\pi\,|_{R(\pi)}$ only.* If $\pi'$ agrees with $\pi$ on $R(\pi)$, the runs on $\pi$ and $\pi'$ coincide step by step: by induction on $t$, $\tau_{\text{ctrl}}$ of the configuration, and its effect is a fixed function of the configuration (the interface clause for calls; Copy, Evict, Const, Move and Emit read no cell) except for a Read, whose result is the block in the head cell, a cell of $R(\pi)$ on which $\pi' = \pi$. *(ii) If $\pi \in D_k$ and $|\, R(\pi) \,| \leq n-k-3$, then $a_{k-1} \in R(\pi)$.* Otherwise $U := [n] \setminus R(\pi)$ has at least $k+3$ elements, so $U \setminus \{a_0 \ldots a_k\}$ contains two distinct cells $y, y'$. Let $\pi' := \pi \circ (a_{k-1}\ y)$ and $\pi'' := \pi \circ (a_{k-1}\ y')$; both are permutations, and both agree with $\pi$ on $R(\pi)$ because they differ from $\pi$ only at $a_{k-1}, y, y' \in U$, so by (i) $\Lambda$ emits the same block on $\pi$, $\pi'$ and $\pi''$. For $i \leq k-2$ the cell $a_i$ lies outside $\{a_{k-1}, y\}$, so $\pi'(a_i) = a_{i+1}$ and, by induction, $\pi'^{\,i}(1) = a_i$ for all $i \leq k-1$; hence $\pi'^{\,k}(1) = \pi'(a_{k-1}) = \pi(y)$, and likewise $\pi''^{\,k}(1) = \pi(y') \neq \pi(y)$. The correct answers on $\pi'$ and $\pi''$ differ while the outputs coincide, contradicting always-correctness. *(iii) Pointwise bound.* The head starts at cell 1 and each Move changes it by $\pm 1$ on a linear tape, so reading cell $x$ costs at least $x-1$ moves, and if $|\, R(\pi) \,| \geq m$ some cell $\geq m$ is read. For $\pi \in D_k$, either $|\, R(\pi) \,| \geq n-k-2$ and $Q_{\text{move}}(\pi) \geq n-k-3$, or $|\, R(\pi) \,| \leq n-k-3$, $a_{k-1} \in R(\pi)$ by (ii) and $Q_{\text{move}}(\pi) \geq a_{k-1} - 1$. *(iv) Expectation.* The length of the cycle of 1 in a uniform $\pi$ is uniform on $[n]$ (the probability of length $\ell$ is $\frac{n-1}{n} \cdot \frac{n-2}{n-1} \cdots \frac{n-\ell+1}{n-\ell+2} \cdot \frac{1}{n-\ell+1} = \frac{1}{n}$), and $D_k$ is the event that this length is at least $k+1$, so $\Pr[D_k] = (n-k)/n \geq 3/4$. Conjugation by a permutation $\tau$ with $\tau(1) = 1$ preserves the uniform measure and $D_k$ and sends $a_{k-1}$ to $\tau(a_{k-1})$, and $a_{k-1} \neq a_0 = 1$ on $D_k$ since $k \geq 2$; so conditioned on $D_k$ the cell $a_{k-1}$ is uniform on $[n] \setminus \{1\}$. Therefore

$$\mathbb{E}_\pi[Q_{\text{move}}] \geq \Pr[D_k] \cdot \frac{1}{n-1} \sum_{a=2}^{n} \min\{a-1, n-k-3\} \geq \frac{3}{4} \cdot \frac{1}{n-1} \sum_{i=1}^{n-k-3} i$$

$$= \frac{3}{4} \cdot \frac{(n-k-3)(n-k-2)}{2(n-1)} \geq \frac{3\,(n-k-3)^2}{8\,n},$$

the last step because $n(n-k-2) \geq (n-1)(n-k-3)$ when $2n \geq k+3$. For $k \leq n/4$ and $n \geq 16$, $n-k-3 \geq 3n/4 - 3 \geq 9n/16$, so $\mathbb{E}_\pi[Q_{\text{move}}] \geq \frac{3}{8} \cdot \frac{81}{256} n = \frac{243}{2048} n > \frac{n}{9}$. □

**Corollary 15** (Zero-error randomisation)**.** *Under the tape and input hypotheses of Proposition 14, every randomised* stream lam *that, on every permutation, terminates and emits the correct block with probability one over its random tape satisfies $\mathbb{E}_{\pi,\omega}[Q_{\text{move}}] \geq \frac{243}{2048} n \geq n/9$; no restriction on $C$, the control, the tags, the counters, the block width or the calls is needed.*

*Proof.* There are finitely many permutations, so the intersection over them of the probability-one events "the run terminates and emits the correct block" is a probability-one set $\Omega_0$ of random tapes. Fix $\omega \in \Omega_0$. With the tape fixed, the operation chosen at each step is a function of the configuration and of the tape bits consumed so far, so the machine is a deterministic always-correct machine of Proposition 14: step (i) of its proof holds because runs on $\pi$ and $\pi'$ that agree on the cells read consume the same tape prefix, and steps (ii)–(iv) use nothing else. Hence $\mathbb{E}_\pi[Q_{\text{move}} \mid \omega] \geq \frac{243}{2048} n$ for every $\omega \in \Omega_0$, and averaging over $\omega$ gives the bound. □

## C.2. Query bandwidth: adaptive observations (Theorem 16)

We now formalize the observation bandwidth available to an adaptive lam execution.

**Query bandwidth.** Consider a task $f : \mathscr{B}_n^m \to \mathscr{B}_n$ with $m$ input blocks. We use $D(f)$ for the minimum worst-case number of input blocks required by a deterministic procedure to compute $f$, and $R_{1/3}(f)$ for the minimum worst-input *expected* number of block queries among randomized procedures that err with probability at most 1/3 on every input. In a LAM execution, input blocks are observed through model-call inputs or outside them. Let $N$ denote the number of model calls, and let $Q_{\text{obs}}$ count every other event that exposes an input block: an address decoded from it, its use as a tool or verifier argument (counted per block), and its emission as an output block (Definition 3). Since the insertion convention gives each model call at most $C-1$ argument blocks, these channels together expose at most $N(C-1)+Q_{\text{obs}}$ input blocks.

**Theorem 16** (Query bandwidth). *Under value provenance and $C \geq 2$, every deterministic LAM computing $f$, its output being the block it EMITS or the tag it halts with, has, on some input, $\sum_{\text{calls}}$(input-block arguments) $+ Q_{\text{obs}} \geq D(f)$, hence $N(C-1)+Q_{\text{obs}} \geq D(f)$. Every randomized LAM with error probability at most 1/3 on each input satisfies $\max_x \mathbb{E}_x[N(C-1)+Q_{\text{obs}}] \geq R_{1/3}(f)$.*

The bound tracks the information the task requires, not how many input blocks could affect the output: for the selector $f(s, d_1, \ldots, d_m) = d_s$ an execution decodes $s$ into an address and emits $d_s$, two observations, and $D(f) = 2$, whereas parity of $m$ blocks needs all $m$, $D(f) = m$. Output is charged through EMIT: identity on a block already in $M$ costs $Q_{\text{obs}} = 1 = D(f)$, whereas a final-configuration convention without such a charge would let it cost nothing, so the theorem is not stated for memory-initialized vertex-granular $G$-evaluations.

**Remark 4** (Content-oblivious schedules). Call a source *globally relevant* if some pair of inputs differing only there changes $f$, and let $r(f)$ be the number of such sources. A LAM with $C \geq 3$ whose schedule does not depend on block contents and whose output is a single block (emitted or terminal, not a halting tag), counting every value-producing call (model or tool) as a call, needs $N \geq (r(f)-1)/(C-2)$ calls, and this is tight for group sums. The output clause is needed: a schedule that loads $m \leq C-1$ blocks and halts with the tag of a parity verifier on them makes no value-producing call. This bound concerns the call complexity of content-oblivious schedules, whereas Theorem 16 bounds the observation budget of arbitrary, adaptive ones; neither dominates the other pointwise, and adaptive access can be far cheaper than any content-oblivious schedule.

*Proof of Remark 4.* For a content-oblivious controller, the structural occurrence DAG — calls, argument slots, addresses and designated output — is the same for every input. Let $o$ be its output occurrence. If a globally relevant source $x_i$ were not an ancestor of $o$, changing only $x_i$ would leave all ancestors and hence the output unchanged for every pair of inputs, contradicting relevance. Thus all $r(f)$ relevant sources occur in the ancestor subgraph.

Every call has at most $C-1$ arguments under the insertion convention. In the ancestor subgraph of $o$, every vertex is an ancestor of the single sink, so every non-sink vertex has out-degree at least one; with $N$ internal vertices of in-degree at most $C-1$ and $r$ sources it therefore has at least $N+r-1$ and at most $N(C-1)$ edges, whence $r \leq N(C-2)+1$. Hence $r(f) \leq N(C-2)+1$ and $N \geq (r(f)-1)/(C-2)$.

For tightness, let $(\mathscr{B}, \oplus)$ be an abelian group and $f = x_1 \oplus \cdots \oplus x_n$. A fixed $(C-1)$-ary fold uses $\lceil (n-1)/(C-2) \rceil$ calls, up to the final partially filled call. □

*Proof of Theorem 16, deterministic case.* Simulate the deterministic LAM by a decision tree over the $m$ input coordinates. Run the machine symbolically: each block carries its provenance label (Definition 3), and an input block whose coordinate has not been queried is kept as a symbol. The transition function sees only $(q, \gamma, a)$, the counters and the occupancy pattern, so the symbolic run proceeds without any block content until an event needs one. There are four such events. A model call: the tree queries the coordinates of its input-block arguments, at most $C-1$ per call, counted through $N$. A SETADDR on an input block, a tool or verifier call with an input block among its arguments, and an EMIT of an input block: each is charged to $Q_{\text{obs}}$, per event and per input block, and the tree queries the coordinate if it has not done so already. By the interface clause of Definition 3 a returned block or tag is a fixed function of the opcode and the argument blocks, and an address set by SETADDR or SHIFT is a fixed function of a block or of the previous address, while a STREAM head position is a function of the controller's own MOVES; so returned blocks, tags and addresses are functions of the queried coordinates and the constants, every subsequent controller state is determined, and at the end every emitted block is a known function of the queried coordinates, as is the tag register at a halting step, so the bound covers a tag output verbatim; the tree outputs $f(x)$. The clause is where the argument would otherwise fail: a verifier with side access to the input could report the parity of $m$ blocks as a tag, for $N = Q_{\text{obs}} = 0$ against $D(f) = m$. On input $x$ it makes at most $N(x)(C-1)+Q_{\text{obs}}(x)$ queries; more precisely, the simulation charges each model call by its number of input-block arguments, so it proves, on some input, $\sum_{\text{calls}}$(input-block arguments) $+ Q_{\text{obs}} \geq D(f)$, of which the statement is the corollary obtained from the bound $C-1$ on the arguments of a call. Since every deterministic decision tree computing $f$ has worst-case depth at least $D(f)$, the claimed worst-case input exists. The output charge is not optional: without it the machine that reads $x_1$ into a slot and emits it computes the identity with $N = Q_{\text{obs}} = 0$ while $D = 1$, and the per-block charge on tool calls is what stops one tool call from absorbing $C-1$ observations for the price of one. *Output semantics.* The theorem's machine computes $f$ by emitting $f(x)$; a block that merely resides in $X$ or $M$ when the machine halts is not an output. Under the final-configuration convention of memory-initialized vertex-granular $G$-evaluations (Section 4) the statement would need an output-observation rule, a terminal exposure of each designated output block charged to $Q_{\text{obs}}$: with it, identity on a source block held in $M$ costs $Q_{\text{obs}} = 1 = D(f)$ under both conventions, since the final exposure of that block is charged; without it, designating the resident source as the output costs nothing and the bound fails. We state the theorem for emitted output and claim it for terminal residency only with that rule. □

*Proof of Theorem 16, randomized case.* Fix the random tape of the LAM; the deterministic construction above gives a decision tree with exactly the same output and, on each input $x$, at most $N(x)(C-1) + Q_{\text{obs}}(x)$ queries. Randomising over tapes therefore gives a randomised decision tree with the same worst-input error and, on each $x$, expected query count at most $\mathbb{E}_x[N(C-1) + Q_{\text{obs}}]$. By the definition of $R_{1/3}(f)$, its worst-input expected count is at least $R_{1/3}(f)$. □

**Remark 5** (Global relevance versus query complexity)**.** Every data coordinate of the selector $f(s, d_1, \ldots, d_m) = d_s$ is globally relevant, but an execution observes only $s$ (one decoded address) and $d_s$ (one emitted input block): $r(f) = m + 1$ while $D(f) = 2$ in the block-query model. The former controls only an input-independent schedule, and the latter is the right quantity for adaptive content addressing.

### C.3. Exact single-stack FIFO frontier

For FIFO retrieval, a single stack exhibits an exact context threshold: $C = \lceil (n+1)/2 \rceil$ with unrestricted buffering during the write phase, and $C = n$ under copy-free streaming.

**Theorem 17** (A single stack needs context $\lceil (n+1)/2 \rceil$ for FIFO)**.** *In* $\text{FIFO}_n$ *the input blocks* $x_1, \ldots, x_n$ *are delivered one at a time, in order, during a* write phase*; a* read phase *begins after* $x_n$ *has been delivered, and during it the machine must emit* $x_1, \ldots, x_n$ *in that order. Nothing is emitted during the write phase. Consider the one-*stack LAM*s of Definition 3 that make no model or tool call (FIFO is a routing task; verifier calls, which return tags only, are allowed, and the finite control, tag and address state may be arbitrarily large), with constant set* $K_0$*, and assume* $|\mathscr{B}| \geq n + |K_0|$*, so that an input of pairwise distinct blocks outside* $K_0$ *exists.*

1. *A* ram-*LAM with* $S \geq n$ *solves* $\text{FIFO}_n$ *with* $C = 1$ *and* $Q_{\text{blk}} = Q = 2n$*.*
2. *Every correct one-*stack *LAM has* $C \geq \lceil (n+1)/2 \rceil$*, and* $C = \lceil (n+1)/2 \rceil$ *suffices (stack capacity* $S \geq C$*), with a fixed content-blind schedule.*
3. *If the write phase is* copy-free streaming*, that is, every block is pushed from its delivery slot before the next delivery, no* Copy *is applied to an input block during the write phase, and nothing is popped before the read phase, then* $C \geq n$*, and* $C = n$ *suffices (with* $S \geq n$*).*

The phase restrictions are necessary for the stated frontier. Without the requirement that the read phase follow the whole write phase, passing each block through solves the task at $C = 1$. Without the clause that nothing is popped in (iii), a push followed by a pop emulates buffering and (iii) collapses to (ii). Without the copy-free clause it collapses likewise: a machine may push every arrival and still retain copies of $x_1, \ldots, x_{C-1}$ in context (at the delivery of $x_i$, $i \leq C-1$, two slots are free, one for the arrival and one for its copy), emit those first in the read phase, and pop $x_n, \ldots, x_C$ into the $C$ slots so freed, which works whenever $n \leq 2C - 1$; so streaming with retained copies costs exactly $\lceil (n+1)/2 \rceil$, the buffered schedule of (ii) in disguise, and (iii) is the price of streaming without them. The bounds do not depend on the controller's content-dependent state: under value provenance that state can choose the next routing operation but can neither hold a block nor regenerate one, so the argument below is pointwise on a single run.

**Lemma 18** (Single-copy normal form)**.** *Fix a machine as in Theorem 17, a capacity* $C$*, and an input of pairwise distinct blocks outside* $K_0$*. From a correct run one obtains a sequence of* Input*,* Push*,* Pop*,* Evict *and* Emit *operations that is legal for the same* $C$*, emits* $x_1, \ldots, x_n$ *in order after all deliveries, and in which at every instant each* $x_i$ *has at most one copy, in a context slot or at one stack position, with* Emit *removing the emitted copy.*

*Proof.* Every block in the run is a copy of some $x_i$ or of a constant, because Input and Const are the only block-creating operations available to the machine. The block emitted at position $i$ has content $x_i$, which is not a constant and differs from every $x_j$, $j \neq i$; so it is a copy of $x_i$. Trace that copy backwards: a copy in a slot arrived by Input, by Pop of a stack position that an earlier Push filled from a slot, or by Copy from a slot; continuing, the trace ends at the delivery of $x_i$. Call the copies on this trace the lineage of $x_i$. Delete every operation acting on a copy outside a lineage and delete every Copy on a lineage and let its source slot continue that lineage, exchanging the names of the source and destination slots in the slot map used for all subsequent operations. The destination slot was free at the Copy, so the exchange is bookkeeping in the construction of the reduced sequence, not a machine operation: after it, every slot named by the original run is either free or holds the block the original run holds there, so the reduced run's occupancy is at most the original's at every instant and no lineage block is ever in two slots. Hence the capacity $C$ and every free-slot precondition are respected. The stack at any instant becomes the subsequence of the original stack formed by lineage positions, so a lineage Pop, which popped the original top, pops the top of the reduced stack. Deliveries and emissions are unchanged, and an emitted copy has no later use on its lineage, so it may be discarded. Lineages of distinct blocks are disjoint and each lineage is a path, so each $x_i$ has at most one copy at any instant. □

*Proof of Theorem 17. (i).* The RAM controller writes each delivered block to address $i$, taken from a public counter, and evicts it; in the read phase it reads address $i$ into its one slot and emits. This uses $C = 1$ and $2n$ transfers.

*(ii), lower bound.* Fix a correct machine and the input of the lemma, and work with the single-copy sequence, in which at every instant each $x_i$ is in a slot, at one stack position, or already emitted. Let $T_w$ be the end of the write phase and let the stack at $T_w$ hold $b_1, ..., b_m$ from bottom to top. If $m = 0$, all $n$ blocks are in slots at $T_w$, since nothing has been emitted, so $n \le C \le 2C - 1$. Otherwise let $t_1$ be the instant of the last Push of $b_1$ before $T_w$ and $t_2$ the instant of the first Pop of $b_1$ after $T_w$. Then $b_1$ is in the stack throughout $[t_1, t_2]$: it is in the stack at $T_w$ and $t_1$ is its last push before $T_w$, so it is not popped in $(t_1, T_w]$, and $t_2$ is its first pop after $T_w$. It occupies the bottom position at $T_w$, and the positions below a resident block never change; so the stack is empty just before $t_1$, and just before $t_2$ it consists of $b_1$ alone. Write $\mathrm{idx}(x_i) = i$ and split the other $n - 1$ blocks into $A_< = \{x : \mathrm{idx}(x) < \mathrm{idx}(b_1)\}$ and $A_> = \{x : \mathrm{idx}(x) > \mathrm{idx}(b_1)\}$.

Just before $t_1$, every block of $A_<$ has been delivered, because deliveries are in index order and $b_1$ has been delivered; none has been emitted, this being the write phase; and none is in the stack, which is empty. So the blocks of $A_< \cup \{b_1\}$ occupy distinct slots, and $|A_<| + 1 \le C$.

Just after $t_2$, $b_1$ has been popped into a slot and has not been emitted, having been in the stack since $t_1$; emissions are in index order, so no block of $A_>$ has been emitted; and the stack is empty. So the blocks of $A_> \cup \{b_1\}$ occupy distinct slots, and $|A_>| + 1 \le C$.

Adding the two counts, $n + 1 = |A_<| + |A_>| + 2 \le 2C$, that is, $C \ge (n + 1)/2$.

*(ii), upper bound.* Let $n \le 2C - 1$; if $n \le C$, keep every block in context. Otherwise deliver $x_1, ..., x_C$ into the $C$ slots and push them in the order $x_C, ..., x_1$, so that $x_1$ is on top and the context is empty; deliver $x_{C+1}, ..., x_n$ into $n - C \le C - 1$ slots. In the read phase pop $x_1$ into the free slot and emit it, then $x_2$, and so on up to $x_C$, and finally emit the buffered blocks in order. The schedule is fixed in advance and inspects nothing.

*(iii).* Argue on the run itself, not on the single-copy sequence: deleting a push would violate the streaming clause. Under copy-free streaming each $x_i$ has a single copy at $T_w$; for $i < n$ it was pushed before $x_{i+1}$ was delivered and never popped, so at $T_w$ the stack holds $x_1, ..., x_{n-1}$ in index order from the bottom, constant blocks possibly interleaved, and $x_n$ is on the stack above them or still in its delivery slot; the context holds no other input block. The first emission is of $x_1$ and needs a copy of $x_1$ in a slot, hence a pop of $x_1$; let $t_2$ be the first one. Just before $t_2$ nothing is above $x_1$ on the stack, so every copy of $x_2, ..., x_n$, whether original or made after $T_w$, is off the stack, popped and not pushed back or never pushed; none of these blocks has been emitted, since $x_1$ has not; and each still has a copy, by value provenance, which is therefore in a context slot. These $n - 1$ slots and the free slot that receives $x_1$ at $t_2$ give $n \le C$. Conversely, with $C = n$ the read phase pops all $n$ blocks into the $n$ slots and emits them in order. The copy-free schedule uses $Q_{\mathrm{blk}} = Q = 2n$ stack operations; the buffered schedule of (ii) uses $2C$ when $C < n \le 2C - 1$ and none when $n \le C$. The storage-optimal construction, with $S \ge (n - C)_+$ and $2(n - C)_+$ operations, is Corollary 19. □

**Remark 6** (The buffered upper bound is tight)**.** Buffering arrivals before pushing them, or retaining copies of the first $C - 1$, halves the copy-free streaming requirement, which is why Theorem 17 states the two regimes separately: $C = n$ against $C = \lceil (n + 1)/2 \rceil$. The lower bound of (ii) counts the context at two instants, the last push and the first pop of the bottom block: every block of smaller index is resident at the first, every block of larger index at the second, and the bottom block at both. The buffered schedule, whose bottom block is $x_C$, meets both counts with equality ($x_1, ..., x_C$ resident before the first push; $x_C, ..., x_{2C-1}$ resident after its pop), so no schedule, content-dependent or not, does better. This pins the requirement at $\lceil (n + 1)/2 \rceil$ blocks, against $C = 1$ for ram. Two stacks would remove it (push each arrival on the first; in the read phase pop all $n$ onto the second and pop them again, emitting: $C = 1$ and $4n$ operations), so the separation is between one stack and either two stacks or random access, not between stacks and random access.

**Corollary 19** (Exact context–stack–traffic trade-off)**.** *Under the hypotheses of Theorem 17, a one-STACK LAM with context $C$ and stack capacity $S$ solves* $\mathrm{FIFO}_n$ *if and only if $C \ge \lceil (n + 1)/2 \rceil$ and $S \ge \max\{n - C, 0\}$, and every such machine makes at least* $2 \max\{n - C, 0\}$ *pushes and pops; a content-blind schedule attains both bounds.*

*Proof.* Necessity of $C \ge \lceil (n + 1)/2 \rceil$ is Theorem 17(ii). Nothing is emitted during the write phase and no block is manufactured (no model or tool call; constants lie in $K_0$, disjoint from the input), so at the end of the write phase every $x_i$ has a copy in a slot or on the stack, and at most $C$ of the $n$ blocks have a copy in a slot; hence at least $m := \max\{n - C, 0\}$ blocks are on the stack only, $S \ge m$, and each of them was pushed at least once and must be popped before it is emitted, giving at least $2m$ pushes and pops. For sufficiency let $m > 0$, so that $m \le C - 1$: keep $x_1$ in a slot, buffer $x_2, ..., x_{m+1}$ as they arrive ($m + 1 \le C$ slots), push them in the order $x_{m+1}, ..., x_2$, which leaves $x_2$ on top, and buffer $x_{m+2}, ..., x_n$, which together with $x_1$ occupy $n - m = C$ slots. In the read phase emit $x_1$, then pop and emit $x_2, \dots, x_{m+1}$ in turn, then emit $x_{m+2}, \dots, x_n$: $m$ pushes, $m$ pops, stack height $m$, and no block content is read. □

### C.4. Proof of Theorem 3

*Proof.* The bit-reversal graph $\Delta_n^{\mathrm{rev}}$ has two rows $u_0, \dots, u_{n-1}$ and $w_0, \dots, w_{n-1}$, path edges along each row, and crossing edges $(u_i, w_{\mathrm{rev}\,(i)})$. It has one source $u_0$, one sink $w_{n-1}$, $2n$ vertices and maximum in-degree two. The $S \geq n$ machine of the last claim is a memory-initialized vertex-granular $G$-evaluation and the machines of the $(C, S)$ law are input-delivered vertex-granular $G$-evaluations (Section 4): all compute each vertex by one call on blocks standing for its predecessors, use no Const, and end with the sink resident or stored; the latter hold no source in $M$ at the start and receive $u_0$ exactly once by Input.

*Persistent memory.* Let $S \geq n$. A topological RAM schedule computes every vertex once, keeps the current $u_i$ or $w_j$ resident, stores each $u_i$ after computing it and loads $u_{\mathrm{rev}\,(j)}$ before computing $w_j$: $N = 2n - 1$, $Q_{\mathrm{blk}} = 2n$, $C = 4$. This is a uniform family-level schedule: public structural counters enumerate the rows, and the fixed permutation $\mathrm{rev}\,(i)$ is applied only to the public index $i$; no block content is inspected.

*Lower bound for capacities* $(C, S)$. The input stream consists of the single source block, delivered once by Input; the source, being an input rather than a call output, cannot be regenerated once every copy of it is gone, so a correct run keeps a copy in $X$ or $M$ until the last call that reads it; the theorem needs no residency hypothesis. Map the run to the black pebble game by placing a pebble on a vertex exactly while some block standing for it (by provenance) is in $X$ or $M$. At most $C + S$ vertices are pebbled at any instant. A vertex becomes pebbled only through the Input, which places the source once, or through the call computing it, whose arguments, blocks for its predecessors, are resident at that moment; loads, stores and slot copies move no pebble, and a pebble is removed when the last copy is evicted. This is the event mapping of Theorem 1(i) with the two colours merged (Remark 1), and it yields a legal pebbling of the ordinary black pebble game with at most $C + S$ pebbles in which the source is placed exactly once and every other placement is one oracle call. The order $n^2/s$ of the call–space trade-off for bit reversal is classical [19]; we prove the bound we use directly, with an explicit constant, so that no pebble range or move convention is imported.

**Lemma 20** (Bit-reversal trade-off). *Let $n = 2^m$, $2 \leq s \leq n/2$, and let $L$ be the least power of two with $L \geq 2s$. Every pebbling of $\Delta_n^{\mathrm{rev}}$ with at most $s$ black pebbles that places a pebble on every $w_j$, recomputation and repeated placements on the source allowed, makes at least $n^2/(4L) > n^2/(16s)$ placements on the row $u_0, \dots, u_{n-1}$ ($n^2/(8s)$ when $s$ is a power of two).*

*Proof.* Let $\ell(j)$ be the time of the first placement on $w_j$. Since a placement on $w_j$ requires $w_{j-1}$ pebbled, $\ell(0) < \ell(1) < \cdots < \ell(n-1)$. Write $L = 2^a$; since $2s \leq n$ and $n$ is a power of two, $4 \leq L \leq n$, and $L < 4s$. Partition the $w$-row into the $n/L$ windows $W_h = \{w_{hL}, \dots, w_{(h+1)L-1}\}$ and let $I_h$ be the time interval $(\ell(hL), \ell((h+1)L - 1)]$; the intervals are pairwise disjoint because $\ell((h+1)L - 1) < \ell((h+1)L)$.

Fix a window. Two indices $j \neq j'$ in it share their top $m - a$ bits and differ in their low $a$ bits, so $\mathrm{rev}\,(j)$ and $\mathrm{rev}\,(j')$ share their low $m - a$ bits and differ in their top $a$ bits: $|\,\mathrm{rev}\,(j) - \mathrm{rev}\,(j')\,| \geq 2^{m-a} = n/L$. Write $r_1 < \cdots < r_L$ for the values $\mathrm{rev}\,(j)$, $j \in W_h$, set $r_0 = -1$, and let $J_i = (r_{i-1}, r_i]$, $i = 1, \dots, L$, be the resulting disjoint intervals of $u$-indices, so that $|\,J_i\,| \geq n/L$ for $i \geq 2$. Let $A$ be the set of $u$-indices carrying a pebble immediately before the placement at time $\ell(hL)$ ("pebbled at time $\ell(j)$" likewise means immediately before the placement of $w_j$), $|\,A\,| \leq s$, and let $P$ be the set of $u$-indices receiving a placement during $I_h$. Call $i$ *blocked* if $J_i$ meets $A$. The $J_i$ being disjoint, at most $s$ indices are blocked, so at least $L - s - 1 \geq L/4$ indices $i \geq 2$ are unblocked.

Let $i \geq 2$ be unblocked, and let $j \in W_h$ have $\mathrm{rev}\,(j) = r_i$. The index of $\mathrm{rev}\,(hL)$ is blocked, since $u_{\mathrm{rev}\,(hL)}$ carries a pebble at time $\ell(hL)$, so $j \neq hL$ and $\ell(j) \in I_h$; as $r_i \notin A$ while $u_{r_i}$ is pebbled at time $\ell(j)$, the vertex $u_{r_i}$ was placed during $I_h$: $r_i \in P$. Moreover, whenever $x \geq 1$ lies in $P$, the placement on $u_x$ found $u_{x-1}$ pebbled, by a pebble either present at time $\ell(hL)$, so that $x - 1 \in A$, or placed during $I_h$, so that $x - 1 \in P$. Descending from $r_i$, and since $J_i$ contains no element of $A$, all of $J_i$ lies in $P$. Hence $|\,P\,| \geq \sum_{i\ \mathrm{unblocked},\, i \geq 2} |\,J_i\,| \geq (L - s - 1)(n/L) \geq n/4$, and at least $n/4$ placements on the $u$-row fall in $I_h$. Summing over the $n/L$ disjoint intervals gives at least $n^2/(4L) > n^2/(16s)$ placements, and $n^2/(8s)$ when $L = 2s$. □

The induced pebbling of the run uses $s = C + S$ pebbles, and every placement on the $u$-row other than the single Input delivery is a model call. For $C + S \leq n/2$ the lemma gives $N > n^2/(16(C + S)) - 1$; for every $C$ and $S$, each of the $2n - 1$ internal vertices needs a call, $N \geq 2n - 1$. Averaging the two bounds, $N \geq n^2/(32(C + S)) + n - 1 \geq (n^2/(C + S) + n)/32$ for $C + S \leq n/2$, since $n - 1 \geq n/32$; for $C + S > n/2$ one has $n^2/(C + S) < 2n$, so $(n^2/(C + S) + n)/32 < 3n/32 \leq 2n - 1$. Hence $N \geq (n^2/(C + S) + n)/32$ for every $n \geq 2$, $C \geq 2$ and $S \geq 0$, the lower bound needing no hypothesis beyond $C \geq 2$; the upper bound below needs $C \geq 3$ and $C + S \geq 4$.

*Feasibility and the constant.* Placing $w_j$ for $j \geq 1$ needs $w_{j-1}$, $u_{\mathrm{rev}\,(j)}$ and a slot for the result, so $C \geq 3$ is necessary for every $S$; at $S = 0$ and $n \geq 4$, $C = 3$ is insufficient: when $w_1$ is placed the three slots hold $w_0$, $u_{n/2}$ and $w_1$, so the source has been evicted and can never return, and $u_{\mathrm{rev}\,(2)} = u_{n/4}$, which precedes $u_{n/2}$ on its chain, can no longer be produced. With $S \geq 1$ and $C = 3$ the schedule below works with every checkpoint stored, so for $n \geq 4$ an input-delivered vertex-granular $G$ -evaluation exists exactly when $C \geq 3$ and $C + S \geq 4$, the hypothesis of the theorem, and the upper bound below is proved on this whole feasible region ($C \geq 4$ is its threshold at $S = 0$). For Lemma 20's $u$-row placement bound, exhaustive search over pebble configurations gives minimum $u$-row placement counts of 100, 49, 392 and 187 at $(n, s) = (16, 3), (16, 4), (32, 3)$ and $(32, 4)$, against 5.3, 4, 21.3 and 16 from $n^2/(16s)$ and 8, 8, 32, 32 from $n^2/(4L)$. The theorem's $1/32$ call constant follows by combining this placement bound with the mandatory internal-vertex calls (below).

*Upper bound for capacities* $(C, S)$. The bound is matched by an elementary checkpoint schedule, so no strategy is imported. Let $C \geq 3, C + S \geq 4$ and $k = \min\{n, C - 3 + S\} \geq 1$. Take the checkpoints $u_{c_\ell}, c_\ell = \lfloor \ell n/k \rfloor, \ell = 0, \ldots, k - 1$; up to $C - 3$ of them are kept resident in reserved slots, the source $c_0 = 0$ among them when $C \geq 4$, in its slot throughout, and the remaining $\max\{k - C + 3, 0\} \leq S$ are stored in $M$ when reached (at $C = 3$ every checkpoint is stored, the source immediately on delivery). First compute $u_1, \ldots, u_{n-1}$ along the chain with two sliding slots, keeping or storing each checkpoint as it is reached: at most $C - 1$ resident blocks and $n - 1$ calls. Then for $j = 0, \ldots, n - 1$ compute $w_j$: let $\ell$ be the largest index with $c_\ell \leq \text{rev}(j)$, load $u_{c_\ell}$ into a sliding slot if it is stored, recompute $u_{c_\ell+1}, \ldots, u_{\text{rev}(j)}$ from it with the two sliding slots, at most $\lceil n/k \rceil - 1$ calls since $\text{rev}(j) - c_\ell \leq \lceil n/k \rceil - 1$, and call once with arguments $w_{j-1}$ (absent for $j = 0$) and $u_{\text{rev}(j)}$. At the moment a block is placed the resident blocks are the resident checkpoints, at most one sliding block, $w_{j-1}$ and the new block, at most $C$; afterwards $w_{j-1}$ and the sliding block are evicted. The run uses no Const, its traversal is generated from public structural indices, and it ends with $w_{n-1}$ resident. Its call count is at most $n - 1 + n\lceil n/k \rceil$, every non-source placement being one fixed-oracle call, and its traffic is at most $k$ stores and $n$ loads, $Q_{\text{blk}} \leq 2n$. If $C + S - 3 \geq n$ then $k = n$ and $N \leq 2n - 1$; otherwise $\lceil n/k \rceil \leq n/(C + S - 3) + 1 \leq 4n/(C + S) + 1$, using $C + S \geq 4$, so $N \leq 4n^2/(C + S) + 2n - 1 \leq 4(n^2/(C + S) + n)$. With $S = 0$ this is the $k = C - 3$ schedule with $n - 1 + n\lceil n/(C - 3) \rceil$ calls; with $S \geq n$ it stores every $u_i$ and makes $2n - 1$ calls. This is a polynomial call–space trade-off, not an exponential one. □

## D. Reliable computation

Throughout, a chain of $n$ dependent transitions is executed; given a correct input, a call returns the correct block with probability $1 - \varepsilon$; the correct block has probability $p_c$ and every incorrect block probability at most $p_c - \gamma$; errors are absorbing.

**Theorem 21** (Sampling amplification)**.** *Suppose that at most $A$ incorrect alternatives have nonzero probability at each correct state and each has probability at most $p_c - \gamma$. Taking a plurality vote over $m = O(\gamma^{-2} \log(An/\delta))$ independent samples per transition yields overall failure probability at most $\delta$ using $N = O(n\gamma^{-2} \log(An/\delta))$ model calls.*

*Proof of Theorem 21.* Fix a step and draw $m$ i.i.d. samples. Let $\hat{p}_y$ be the empirical frequency of block $y$. By Hoeffding, $\Pr[\hat{p}_y - \hat{p}_{y^\star} \geq 0] \leq \exp(-m\gamma^2/2)$ for each of the at most $A$ incorrect alternatives $y$, since $\mathbb{E}[\hat{p}_{y^\star} - \hat{p}_y] \geq \gamma$ and the difference is an average of $m$ i.i.d. variables in $[-1, 1]$. A union bound over $y$ gives per-step error $\leq A \exp(-m\gamma^2/2)$, which is $\leq \delta/n$ for $m \geq 2\gamma^{-2} \log(An/\delta)$. A union bound over the $n$ steps gives overall failure $\leq \delta$, at $N = nm = O(n\gamma^{-2} \log(An/\delta))$. □

*Proof of Theorem 4. Hard family without a lookahead signal.* Consider the complete binary tree of depth $n$, with a node represented by its root-to-node bit string (so the direct encoding uses $b(n) = \Theta(n)$ bits at maximum depth). Every internal node $u$ carries an independent hidden bit $\theta_u \sim \text{Ber}(1/2)$. A call at $u$ returns child $u\theta_u$ with probability $p_c = \frac{1}{2} + \frac{\gamma}{2}$ and the other child $u(1 - \theta_u)$ with probability $\frac{1}{2} - \frac{\gamma}{2}$. Repeated calls at the same node are conditionally i.i.d. given $\theta_u$. Starting at the root, the designated correct path is obtained by following the preferred child $\theta_u$ at every visited correct-path node.

Crucially, an erroneous commitment does not trigger a special oracle output. The reached off-path node has its own fresh independent $\theta_u$ and exactly the same two-candidate sampling law as every other node. Hence future observations are distributionally ordinary and independent of the hidden bit at the node where the wrong turn occurred. Tree branches never merge, so a wrong commitment is nevertheless absorbing for the final path. This removes the one-step lookahead diagnostic that would arise if wrong prefixes emitted a distinguished failure symbol.

At stage $j$, conditioned on all previous commitments being correct, let $H_{j-1}$ denote the full pre-stage history. The current node $u$ is then fixed by $H_{j-1}$, while $\theta_u$ is still uniform and independent of that history. A stage-local strategy draws a stopping-time number $N_j$ of samples at $u$ before committing. For a fixed history $h$ put

$$a_j(h) = \frac{1}{2}\left(\mathbb{E}_0[N_j \mid H_{j-1} = h, S_{j-1}] + \mathbb{E}_1[N_j \mid H_{j-1} = h, S_{j-1}]\right).$$

Distinguishing the two possible preferred-child laws is exactly testing $\text{Ber}(\frac{1}{2} + \frac{\gamma}{2})$ against $\text{Ber}(\frac{1}{2} - \frac{\gamma}{2})$. Their directed KL divergence is

$$D = \gamma \log \frac{1 + \gamma}{1 - \gamma} \leq 3\gamma^2 \qquad (\gamma \leq \frac{1}{2}).$$

Fix $h$ and let $P_0, P_1$ be the laws of the stopped sample sequence at $u$ under $\theta_u = 0$ and $\theta_u = 1$, with expected lengths $m_0 = \mathbb{E}_0[N_j \mid H_{j-1} = h, S_{j-1}]$ and $m_1 = \mathbb{E}_1[N_j \mid H_{j-1} = h, S_{j-1}]$, so that $a_j(h) = (m_0 + m_1)/2$. If $m_0$ or $m_1$ is infinite then $\mathbb{E}[N] = \infty$ and there is nothing to prove, so let both be finite. Wald's identity [41] gives the two directed divergences of the stopped experiment, $\mathrm{KL}(P_0 \parallel P_1) = D\,m_0$ and $\mathrm{KL}(P_1 \parallel P_0) = D\,m_1$ (the per-sample divergence is $D$ in both directions, the two Bernoulli laws being mirror images). The Bretagnolle–Huber inequality [42], in the form of A. B. Tsybakov [43, Lemma 2.6 and Theorem 2.2], applied to the commitment event $A = \{\hat{\theta}_j = 1\}$ once with $(P_0, P_1)$ and once with the roles exchanged, gives $P_0(A) + P_1(A^c) \geq \frac{1}{2}e^{-Dm_0}$ and $P_0(A) + P_1(A^c) \geq \frac{1}{2}e^{-Dm_1}$, hence $P_0(A) + P_1(A^c) \geq \frac{1}{2}e^{-D\min\{m_0,m_1\}} \geq \frac{1}{2}e^{-Da_j(h)}$, the minimum being at most the average; this is why the average stopped sample count $a_j(h)$ suffices even when the two conditional expected counts differ. Since $\theta_u$ is uniform given $h$, the error probability is half that sum:

$$\Pr[\hat{\theta}_j \neq \theta_u \mid H_{j-1} = h, S_{j-1}] = \frac{1}{2}\left(P_0(A) + P_1(A^c)\right) \geq \frac{1}{4}e^{-Da_j(h)}.$$

Average over $H_{j-1}$ and use Jensen's inequality, since $e^{-Dx}$ is convex. With $a_j = \mathbb{E}[a_j(H_{j-1}) \mid S_{j-1}]$,

$$e_j := \Pr[\hat{\theta}_j \neq \theta_u \mid S_{j-1}] \geq \frac{1}{4}e^{-Da_j}.$$

Let $S_j$ be the event that the first $j$ commitments follow the designated correct path. The events are nested and overall success is $S_n$. Therefore

$$1 - \delta \leq \Pr[S_n] = \prod_{j=1}^{n}(1 - e_j) \leq \exp\left(-\frac{1}{4}\sum_{j=1}^{n} e^{-Da_j}\right).$$

For $\delta \leq \frac{1}{2}$, $-\log(1-\delta) \leq 2\delta$, so $\sum_j e^{-Da_j} \leq 8\delta$. By AM–GM,

$$\prod_{j=1}^{n} e^{-Da_j} \leq \left(\frac{1}{n}\sum_{j=1}^{n} e^{-Da_j}\right)^n \leq \left(\frac{8\delta}{n}\right)^n,$$

and hence

$$\sum_{j=1}^{n} a_j \geq \frac{n}{D}\log\frac{n}{8\delta}.$$

Moreover $\Pr[S_{j-1}] \geq \Pr[S_n] \geq 1 - \delta$, and conditioned on $S_{j-1}$ the hidden bit at the current node is uniform, so

$$\mathbb{E}[N] \geq \sum_{j=1}^{n} \mathbb{E}[N_j \mathbf{1}_{S_{j-1}}] = \sum_{j=1}^{n} \Pr[S_{j-1}]\,a_j \geq (1-\delta)\sum_{j=1}^{n} a_j.$$

Using $\delta \leq \frac{1}{4}$ and $D \leq 3\gamma^2$ gives

$$\mathbb{E}[N] \geq \frac{n}{4\gamma^2}\log\frac{n}{8\delta},$$

as claimed. For $n \geq 64\delta$ one has $\log\frac{n}{8\delta} \geq \frac{1}{2}\log\frac{n}{\delta}$, so the bound is at least $\frac{n}{8\gamma^2}\log\frac{n}{\delta}$, the stated order; since $\delta \leq \frac{1}{4}$, every $n \geq 16$ qualifies. For $n \leq 8\delta$ the displayed bound is non-positive and the theorem is vacuous. □

**Remark 7.** The lower bound is intentionally scoped to the stage-local no-feedback class defined in Section 7. Adaptive sample counts may depend on the committed past and all current-stage samples; what is excluded is using later transitions to revise an earlier commitment. The binary-tree construction additionally ensures that later calls carry no distinguished off-path symbol, so an off-path trajectory is locally indistinguishable from the correct one and the restriction hides no lookahead signal.

**Theorem 22** (Perfect local verification)**.** *With a perfect local verifier, fresh independent attempts, $\varepsilon < 1$ and no cap on retries, sampling a transition until the verifier accepts terminates almost surely with zero error and $\mathbb{E}[N_{\text{model}}] = \mathbb{E}[N_{\text{ver}}] = n/(1-\varepsilon)$ and $\mathbb{E}[\text{Cost}] = n(1+c_v)/(1-\varepsilon)$, which is $\Theta(n)$ for fixed $\varepsilon < 1$ and $c_v$. On the hard family of Theorem 4, this removes the $\Theta(\log(n/\delta))$ overhead of stage-local sampling for constant $\gamma$ and $c_v$.*

*Proof of Theorem 22.* With a perfect verifier the agent repeats a step until acceptance. Attempts are geometric with success probability $1-\varepsilon$, so $\mathbb{E}[N_{\text{model}}] = \mathbb{E}[N_{\text{ver}}] = n/(1-\varepsilon)$ and $\mathbb{E}[\text{Cost}] = n(1+c_v)/(1-\varepsilon)$. No wrong state is propagated. On the binary-tree hard family of Theorem 4, $1-\varepsilon = (1+\gamma)/2$, so stage-local sampling-only strategies need a $\Theta(\log(n/\delta))$ factor more at constant $\gamma, c_v$. □

*Proof of Theorem 5.* One attempt falls into three disjoint outcomes: correct and unanimously accepted with probability $a_t = (1-\varepsilon)(1-p_r)^t$, wrong and unanimously accepted with probability $b_t = \varepsilon p_a^t$, or rejected. The first accepted attempt therefore arrives after a geometric number of trials with parameter $s_t = a_t + b_t$ and is wrong with conditional probability $q_t = b_t/s_t$. This is the per-transition law of the theorem, valid at every transition whose earlier accepted transitions are correct; a union bound over the first transition that accepts a wrong candidate gives chain failure at most $nq_t$ without further assumption. Summing over $n$ accepted steps gives the total call counts and weighted cost, and the summation uses that the attempt law is the same at every visited state; under that law the $n$ accepted candidates are wrong independently with probability $q_t$ each, so the chain fails with probability exactly $1-(1-q_t)^n \le nq_t$. If the law may differ at a state reached through a false accept, the same computation applies conditioned on every earlier accepted transition being correct, an event of probability at least $1-nq_t$: the number of rejected attempts at a stage is independent of whether its accepted candidate is correct (competing hazards), so the conditioning changes no attempt count, and on that event every visited state is correct.

For $t = 1$, $nq_1 \le \delta$ rearranges exactly to

$$p_a \le \frac{\delta(1-\varepsilon)(1-p_r)}{\varepsilon(n-\delta)}.$$

For general $t$, accepted-error odds are

$$\frac{q_t}{1-q_t} = \frac{\varepsilon}{1-\varepsilon}\left(\frac{p_a}{1-p_r}\right)^t.$$

Thus unanimous repetition improves accuracy only if $p_a < 1 - p_r$. Since chain failure is at most $nq_t$, the target $nq_t \le \delta$ holds as soon as $q_t/(1-q_t) \le \delta/(n-\delta)$, that is, when $0 < p_a < 1 - p_r$, for every $t \ge \log\left(\frac{\delta(1-\varepsilon)}{(n-\delta)\varepsilon}\right) / \log\left(\frac{p_a}{1-p_r}\right)$ (one check suffices when $p_a = 0$). Because $s_t \ge a_t = (1-\varepsilon)(1-p_r)^t$ can itself shrink exponentially, the exact cost retains the false-reject penalty that an $O(ntc_v)$ approximation would omit. □

**Remark 8** (Short-circuit checking)**.** If the $t$ checks of a candidate are run in sequence and stopped at its first rejection, the acceptance probabilities $a_t$ and $b_t$, hence $s_t$, $q_t$ and the attempt and model-call counts of Theorem 5, are unchanged; only the number of checks per attempt changes. A correct candidate is checked $\sum_{j=0}^{t-1}(1-p_r)^j$ times in expectation and a wrong one $\sum_{j=0}^{t-1} p_a^j$ times, so an attempt makes $v_t = (1-\varepsilon)\sum_{j<t}(1-p_r)^j + \varepsilon\sum_{j<t} p_a^j \le t$ checks in expectation. Attempts are i.i.d. and the number of attempts is a stopping time with mean $1/s_t$, so Wald's identity [41] gives $\mathbb{E}[N_{\text{ver}}] = v_t/s_t$ per accepted transition and $\mathbb{E}[\text{Cost}] = (1 + c_v v_t)/s_t$ in place of $(1 + tc_v)/s_t$. With at most $M$ attempts per transition, the last candidate being kept if all are rejected (Corollary 25(ii)), attempt $j \le M$ is reached with probability $(1-s_t)^{j-1}$, an event determined by the earlier attempts and independent of the $j$-th attempt's checking cost, so it contributes $(1-s_t)^{j-1}(1 + c_v v_t)$ in expectation; summing over $j \le M$ and the $n$ executed transitions gives $\mathbb{E}[\text{Cost}] = n(1 + c_v v_t)\left(1-(1-s_t)^M\right)/s_t$, the forced commitment at the cap changing correctness but not this cost sum. For $p_r > 0$ and fixed $(\varepsilon, p_a, p_r)$, $v_t \to (1-\varepsilon)/p_r + \varepsilon/(1-p_a)$ as $t \to \infty$, so the factor $1 + c_v\Theta(\log(n/\delta))$ of Corollary 25 is a cost of paying for all $t$ checks and disappears, whereas the $n^{1+\alpha}$ false-reject penalty in model calls, which comes from $s_t$, remains; Theorem 5 as stated is the conservative variant. If verifier rates depend on the candidate (a correct candidate is accepted with a random probability $U$, a wrong one with $V$, checks independent given the candidate), the powers become moments, $a_t = (1-\varepsilon)\mathbb{E}[U^t]$ and $b_t = \varepsilon\mathbb{E}[V^t]$, and the ratio $b_t/a_t$ need not fall geometrically (Corollary 23).

**Corollary 23** (Candidate-dependent checking)**.** *Suppose candidate attempts are i.i.d., a candidate is correct with probability $1-\varepsilon$, and, conditional on the candidate, its $t \ge 1$ checks are independent, each accepting a correct candidate with a candidate-specific probability $U$ and a wrong one with probability $V$ ($U$, $V$ random, drawn with the candidate). Put $a_t = (1-\varepsilon)\mathbb{E}[U^t]$, $b_t = \varepsilon\mathbb{E}[V^t]$ and $s_t = a_t + b_t > 0$. Then retry-until-accept with $t$ unanimous checks commits a wrong candidate with probability $q_t = b_t/s_t$ and makes $1/s_t$ model calls and $t/s_t$ verifier calls in expectation per accepted step, and the chain-level conclusions of Theorem 5 (*Pr[chain failure] $\le nq_t$*; exactly $1-(1-q_t)^n$ with totals $n/s_t$, $nt/s_t$, $n(1+tc_v)/s_t$ under the stationary law) hold verbatim. The odds $q_t/(1-q_t) = \varepsilon\,\mathbb{E}[V^t]/((1-\varepsilon)\mathbb{E}[U^t])$ need not fall geometrically in $t$; Theorem 5 is the case $U \equiv 1-p_r$, $V \equiv p_a$.*

*Proof.* Conditional on a correct candidate with acceptance probability $U$, its $t$ independent checks all accept with probability $U^t$; averaging over $U$ gives Pr[attempt correct and accepted] $= a_t$, and likewise Pr[attempt wrong and accepted] $= b_t$. Attempts being i.i.d., attempt $j$ is the first accepted one and is wrong with probability $(1-s_t)^{j-1}b_t$; summing over $j \ge 1$ gives $b_t/s_t$. The number of attempts is geometric with mean $1/s_t$, and each attempt costs one model call and $t$ verifier calls. The union bound over the first error and, under the stationary law, the product $(1-q_t)^n$ and the summed totals are the calculations of the proof of Theorem 5 with $a_t, b_t$ in place of $(1-\varepsilon)(1-p_r)^t$, $\varepsilon p_a^t$. The odds are $b_t/a_t$ by definition of $q_t$. □

**Corollary 24** (Batch-acceptance account)**.** *Fix a checking procedure that makes $t$ checks per candidate and accepts on an event $A_t$. Attempts, including candidate generation and the checking randomness, are i.i.d.; the checks within one attempt may be* *arbitrarily dependent. Let $a_t = \Pr[\text{candidate correct}, A_t]$, $b_t = \Pr[\text{candidate wrong}, A_t]$ and $s_t = a_t + b_t > 0$. Then retry-until-accept commits a wrong candidate with probability $q_t = b_t/s_t$ and makes $1/s_t$ model calls and $t/s_t$ verifier calls in expectation per accepted transition, and the chain-level conclusions of Theorem 5 hold verbatim with these $a_t, b_t$ under its stationary trajectory law and absorbing-error objective. Theorem 5 is the case $A_t = \{\text{all } t \text{ checks accept}\}$ with conditionally independent checks and Corollary 23 the case with candidate-specific rates.*

*Proof.* The proof of Corollary 23 uses only that attempts are i.i.d. with $\Pr[\text{correct and accepted}] = a_t$ and $\Pr[\text{wrong and accepted}] = b_t$: attempt $j$ is the first accepted one and is wrong with probability $(1-s_t)^{j-1} b_t$, which sums to $b_t/s_t$; the number of attempts has tail $(1-s_t)^j$, hence mean $1/s_t$; and every attempt makes one model call and $t$ verifier calls. The union bound, the product $(1-q_t)^n$ and the summed totals follow as there. No independence among the checks of one attempt enters. □

**Corollary 25** (False-reject penalty; the capped policy). *(i) Let $0 < p_a < 1 - p_r < 1$, $0 < \varepsilon < 1$, $0 < \delta < 1$ and $n \geq 2$, put $\rho = p_a/(1-p_r)$, $\mathscr{K}_{\text{pen}} = \big(\frac{(n-\delta)\varepsilon}{\delta(1-\varepsilon)}\big)^\alpha$ with*

$$\alpha = \frac{-\log(1-p_r)}{\log((1-p_r)/p_a)} > 0,$$

*assume the stationary attempt law of Theorem 5 at every visited state, and let $t$ be the smallest integer with $nq_t \leq \delta$, which is at least 1 exactly when $n\varepsilon > \delta$, assumed here. Then*

$$\frac{(n-\delta)\mathscr{K}_{\text{pen}}}{1-\varepsilon} \leq \mathbb{E}[N_{\text{model}}] = \frac{n}{s_t} \leq \frac{n\mathscr{K}_{\text{pen}}}{(1-\varepsilon)(1-p_r)},$$

*with $\mathbb{E}[\text{Cost}] = (1+tc_v)\,\mathbb{E}[N_{\text{model}}]$ and $t =$ ceil.l$\log \mathscr{K}_{\text{pen}}^{1/\alpha}/\log(1/\rho)$ceil.r. For fixed $(\varepsilon, p_a, p_r)$ the model-call count is therefore $\Theta\big(n(n/\delta)^\alpha\big)$, polynomial rather than linear in $n/\delta$, and the weighted cost carries a further factor $1 + c_v\,\Theta(\log(n/\delta))$; the linear scaling of Theorem 22 is recovered only when $\alpha\log(n/\delta) = O(1)$, i.e. at fixed $p_a$, when $p_r \to 0$ fast enough that $\alpha = O(1/\log(n/\delta))$; for fixed $p_a$ and fixed $p_r > 0$ the count is superlinear. Finally $\alpha < 1$ exactly when $p_a < (1-p_r)^2$. (ii) With $t = 1$, $s = s_1$, $r = 1 - s$ and at most $M$ attempts, the last candidate being kept if all are rejected, the expected numbers of attempts, model calls and verifier calls are each $(1-r^M)/s$, and the committed candidate (the accepted one, or the last one kept by the cap) is incorrect with probability $b_1(1-r^M)/s + r^{M-1}\varepsilon(1-p_a)$, the second term being the price of the cap, so that a chain of $n$ transitions commits $n\big[\,b_1(1-r^M)/s + r^{M-1}\varepsilon(1-p_a)\,\big]$ wrong states in expectation under the stationary law.*

*Proof.* (i) By the proof above, $nq_t \leq \delta$ holds if and only if $\frac{\varepsilon}{1-\varepsilon}\rho^t \leq \frac{\delta}{n-\delta}$, i.e. $t \geq t^\star := \log \mathscr{K}_{\text{pen}}^{1/\alpha}/\log(1/\rho)$, so the smallest sufficient $t$ is $\lceil t^\star \rceil$ and $t^\star \leq t < t^\star + 1$; here $t^\star > 0$, equivalently $\mathscr{K}_{\text{pen}} > 1$, exactly when $(n-\delta)\varepsilon > \delta(1-\varepsilon)$, i.e. $n\varepsilon > \delta$. Since $(1-p_r)^{-t^\star} = \exp\big(t^\star \log\frac{1}{1-p_r}\big) = \mathscr{K}_{\text{pen}}$ by the definition of $\alpha$, we have $\mathscr{K}_{\text{pen}} \leq (1-p_r)^{-t} < \mathscr{K}_{\text{pen}}/(1-p_r)$. The upper bound uses $s_t \geq a_t = (1-\varepsilon)(1-p_r)^t$. For the lower bound, at $t \geq t^\star$ the accepted-error odds $b_t/a_t$ are at most $\delta/(n-\delta)$, so $s_t = a_t(1 + b_t/a_t) \leq a_t n/(n-\delta)$ and $n/s_t \geq (n-\delta)/a_t \geq (n-\delta)\mathscr{K}_{\text{pen}}/(1-\varepsilon)$; the two sides of the bracket differ by the factor $n/((n-\delta)(1-p_r))$. The cost identity is $\mathbb{E}[\text{Cost}] = n(1+tc_v)/s_t$, and $t = \Theta(\log(n/\delta))$ for fixed parameters. Finally $\alpha < 1$ iff $-\log(1-p_r) < \log((1-p_r)/p_a)$ iff $p_a < (1-p_r)^2$, and $\alpha \to 0$ as $p_r \to 0$. (ii) Attempt $j \leq M$ is reached with probability $r^{j-1}$, so the expected number of attempts is $\sum_{j=1}^M r^{j-1} = (1-r^M)/s$, each attempt costing one model call and one verifier call. The accepted candidate is wrong either because some attempt $j \leq M$ was wrongly accepted, probability $\sum_{j\leq M} r^{j-1} b_1 = b_1(1-r^M)/s$, or because all $M$ attempts were rejected and the $M$-th, kept by the cap, is wrong, probability $r^{M-1} \cdot \varepsilon(1-p_a)$, the last attempt being rejected-and-wrong with probability $\varepsilon(1-p_a)$. □

*Proof of Theorem 6. Exact cost.* Assume first that $k \mid n$. A segment of $k$ transitions plus one verification costs $k + c_v$ and is accepted iff all $k$ transitions are correct, which has probability $(1-\varepsilon)^k$. Attempts are i.i.d., so the expected number of attempts is $(1-\varepsilon)^{-k}$ and the expected cost of a segment is $(k+c_v)(1-\varepsilon)^{-k}$. There are $n/k$ segments and rollback is to the last accepted checkpoint, so segments are independent: $T(k) = \frac{n}{k}(k+c_v)(1-\varepsilon)^{-k}$. If $n = qk + r$ with $0 < r < k$, the exact expression is

$$q(k+c_v)(1-\varepsilon)^{-k} + (r+c_v)(1-\varepsilon)^{-r},$$

so the displayed $T(k)$ is not called exact outside the divisibility assumption.

*Continuous optimum.* For $0 < \varepsilon < 1$, let $\beta = -\log(1-\varepsilon)$ (so $\varepsilon \leq \beta$, with $\beta \leq \varepsilon(1+\varepsilon)$ for $\varepsilon \leq \frac{1}{2}$) and $F(k) = \log\big(T(k)/n\big) = \log(1 + c_v/k) + \beta k$. Then

$$F'(k) = \beta - \frac{c_v}{k(k+c_v)}, \qquad F''(k) = \frac{c_v(2k+c_v)}{k^2(k+c_v)^2} > 0,$$

so $F$ is strictly convex; $F'(0^+) = -\infty$ and $F'(\infty) = \beta > 0$, so $F'$ has a unique zero, solving $k^2 + c_v k = c_v/\beta$, whence $k^\star = \frac{1}{2}(\sqrt{c_v^2 + 4c_v/\beta} - c_v)$.

*Two-sided bound on the optimum.* Put $z = \sqrt{c_v \beta}$. For the lower bound, use $e^y \geq 1 + y$ and then AM–GM:

$$\frac{T(k)}{n} = \left(1 + \frac{c_v}{k}\right)e^{\beta k} \geq \left(1 + \frac{c_v}{k}\right)(1 + \beta k) = 1 + \beta k + \frac{c_v}{k} + c_v\beta \geq 1 + 2z + z^2,$$

uniformly in $k$ (AM–GM on the two middle terms, and $c_v\beta = z^2$), so $T(k^\star)/n \geq 1 + 2z + z^2 \geq 1 + 2\sqrt{c_v\varepsilon}$. For the upper bound evaluate at the feasible point $k_1 = \sqrt{c_v/\beta}$, where $c_v/k_1 = \beta k_1 = z$, giving $T(k^\star)/n \leq T(k_1)/n = (1+z)e^z$. The gap between the two bounds is $(1+z)e^z - (1 + 2z + z^2) = \frac{1}{2}z^2 + O(z^3) = O(c_v\beta)$, so $T(k^\star)/n = 1 + 2\sqrt{c_v\beta} + O(c_v\beta)$. Both bounds hold for every $0 < \varepsilon < 1$ and $c_v > 0$ in the positive-real relaxation, provided the horizon permits $\sqrt{c_v/\beta} \leq n$.

*Leading order and finite horizons.* Since $c_v\beta \to 0$ implies $4c_v/\beta \gg c_v^2$, $k^\star = \sqrt{c_v/\beta}\,(1 + O(\sqrt{c_v\beta}))$. If in addition $\varepsilon \to 0$, then $\beta = \varepsilon(1 + O(\varepsilon))$ and this becomes $k^\star = \sqrt{c_v/\varepsilon}(1 + o(1))$. Since $F = \log(T/n)$ is strictly convex, the minimiser of $T$ over the integers $k \geq 1$ is $\max\{1, \lfloor k^\star \rfloor\}$ or $\lceil k^\star \rceil$; hence the optimal integer checkpoint interval lies within one of the continuous minimiser $k^\star$. Restricting to divisors of $n$ gives a coarser admissible grid. For finite $n$, clip the continuous solution to $[1, n]$ and evaluate the exact divisible or remainder expression over admissible integer $k$. At $\varepsilon = 0$ there are no failed segments; verification may be omitted, or a required final check is minimised by $k = n$. The closed form, rather than its leading-order approximation, should be used unless both $\varepsilon$ and $c_v\varepsilon$ are small.

*Correspondence with Young–Daly.* Writing MTBF $= 1/\varepsilon$ (mean transitions between errors) and checkpoint cost $c_v$, the classical first-order optimum $\sqrt{2 \cdot \text{cost} \cdot \text{MTBF}}$ [27], [28] coincides with $k^\star$ up to the factor $\sqrt{2}$, which arises because the classical model loses on average half a segment per failure while ours re-executes the whole segment. This distinction, and the resulting first-order optimum without the factor 2, are those of the silent-error checkpoint-with-verification model of the resilience literature [44], [45], of which Theorem 6 is the discrete, per-transition instance; for this discrete per-transition model, the calculation above gives the closed-form minimiser, the neighbouring-integer rule and the bracket $1 + 2z + z^2 \leq T(k^\star)/n \leq (1+z)e^z$. □

**Remark 9** (The checkpoint policy as a machine of Definition 3; verifiers whose cost grows with $k$)**.** Section 7 states its results over strategy classes; the policies are machines of Definition 3 once $L$ is stochastic, i.e. consumes fresh random-tape bits at each call, so that (opcode, arguments) determine a distribution over (block, tag) rather than a value. The policy of Theorem 6 runs on ram(1) with $C = 2$. The current state occupies one slot; a transition is one $L$-call on it, the returned block taking the second slot and the old state being evicted. At the start of a segment the state is written to $M[1]$ (the checkpoint, one unit of $Q_{\text{blk}}$); after $k$ transitions the checkpoint is read back into the free slot (one unit) and the verifier is called on the two blocks (checkpoint, current state), returning accept or reject as a tag; on reject the current state is evicted and the checkpoint block becomes the current state, so rollback moves no further block; on accept the checkpoint copy is evicted and the current state is written as the next checkpoint. Under the absorbing-state semantics of Section 7, in which an erroneous transition never returns to the correct trajectory, the endpoint is correct exactly when all $k$ transitions were, so this two-argument verifier is the theorem's perfect segment verifier, and its cost $c_v$ does not depend on $k$. The operation trace is exact: initialisation writes $x_0$ to $M[1]$ (one movement); a rejected attempt costs $k$ model calls, one read, one verifier call and no write; an accepted non-final segment costs $k$ model calls, one read, one verifier call and one write; the final accepted segment omits the write and EMITS the state. Hence $\mathbb{E}[N_{\text{model}} + c_v N_{\text{ver}}] = T(k)$, $Q_{\text{blk}} = 1 + N_{\text{ver}} + (n/k - 1) = n/k + N_{\text{ver}}$ exactly, $T_{\text{in}} \leq b(N_{\text{model}} + 2N_{\text{ver}})$ and $Q_{\text{move}} = 0$; every accepted checkpoint is correct, so $T(k)$ is unconditional. The policy of Theorem 5 is the same machine with $k = 1$, $t$ verifier calls on the two resident blocks (state, candidate) per attempt, eviction of the candidate on any reject and no write, so $N_{\text{ver}} = tN_{\text{model}}$, $Q_{\text{blk}} = 0$ and $C = 2$; every reliability policy of Section 7 is thus a machine of Definition 3. If instead the verifier must see the $k$ intermediate states, it needs $k + 1$ argument blocks resident at once, hence $C \geq k + 1$ (a verifier call inserts no block, Definition 3), and a cost $V(k)$ that grows with $k$: the per-transition objective becomes $(1 + V(k)/k)e^{\beta k}$, which for $V(k) = c_v$ is the theorem and for $V(k) = vk$ equals $(1 + v)e^{\beta k}$, increasing in $k$, so the optimum is $k = 1$, i.e. Theorem 22 with $c_v = v$. For the mixed cost $V(k) = v + \eta k$ it equals $(1 + \eta)\left(1 + \frac{v/(1+\eta)}{k}\right)e^{\beta k}$, the theorem's objective with $c_v$ replaced by $c' = v/(1 + \eta)$ and a constant factor $1 + \eta$, so $k^\star = \frac{1}{2}\left(\sqrt{c'^2 + 4c'/\beta} - c'\right)$: a fixed-plus-proportional verifier keeps the square-root law with the fixed part discounted by $1 + \eta$. The law is specific to verifiers whose cost has a fixed part; a purely proportional cost returns the optimum to $k = 1$.

## E. Extensions and deployment mapping

We record several extensions of the base model and describe how its resource counters map to deployment-level measurements.

**Mapping to serving traces.** A deployment trace records model, verifier, and tool calls together with semantic-memory transfers and logical input/output traffic. Tool calls may invoke domain models far from the language model itself, such as camera-controllable video generators [46], multi-scale flow-matching generators [47], physics-informed neural-operator controllers [48], or perception modules for tracking and neural-field reconstruction [49], [50]; each is charged as an oracle call whose argument and output sizes enter the same counters. Prefix caching introduces a distinct physical-prefill quantity: $T_{\text{in}}^{\text{logical}}$ records the oracle argument, whereas $T_{\text{in}}^{\text{prefill}}$ records tokens whose KV state is actually computed. Runtime KV movement is a separate systems resource and should not be identified with semantic block movement.

A scalar deployment cost is obtained by assigning explicit weights to these events. When event costs depend on call size, cache state, or execution parallelism, the event-level costs should be aggregated directly rather than represented by a single universal coefficient.

**Charged controller computation.** Strict VPA can be relaxed by representing value-producing controller operations as explicit charged provenance nodes. Query-bandwidth bounds remain unchanged after charging source observations, while the pebbling correspondence applies to the resulting augmented occurrence DAG rather than necessarily to the original DAG. Pointer-chasing bounds continue to apply when unseen memory contents cannot be manufactured locally, and reliability bounds require that controller computation provide no uncharged correctness signal. There is no task-independent conversion from a controller-work budget to a universal loss in the original lower bounds: the effect depends on which semantic values the added operations can create.

**Control state and the single-stack frontier.** Theorem 17 is unchanged by arbitrarily rich tags or content-dependent routing state so long as that state cannot store or re-materialize semantic blocks. Allowing it to do so introduces an additional semantic memory and therefore changes the machine class.

**Proposition 26** (Joint local verification and segment rollback)**.** *Use the imperfect local verifier of Theorem 5 with $t \geq 0$ unanimous checks per accepted step ($t = 0$ denoting no local check, with $s_0 = 1$ and $q_0 = \varepsilon$), and let $g_t = (1 + tc_v)/s_t$ and $q_t = b_t/s_t$. After every $k$ accepted steps, call an independent perfect segment verifier of cost $c_s$ that accepts exactly when all $k$ states are correct; on rejection, roll back the segment. Under the theorem's stationarity and independence assumptions, for $k \mid n$ the exact expected weighted cost is*

$$T(t,k) = \frac{n}{k}(kg_t + c_s)(1 - q_t)^{-k}.$$

*For fixed $t$ and $0 < q_t < 1$, put $\beta_t = -\log(1 - q_t)$. The positive-real minimiser is*

$$k_t^\star = \frac{\sqrt{c_s^2 + 4g_t c_s/\beta_t} - c_s}{2g_t},$$

*clipped to $[1, n]$; the exact finite optimum enumerates integer $t, k$ and the shorter final segment. Interior hybrids can strictly beat both no-local-check checkpointing and sufficient local checking without rollback.*

*Proof.* One accepted local step has expected cost $g_t$ and residual error $q_t$. In the competing hazards calculation of Theorem 5, the number of rejected attempts is independent of whether the first accepted attempt is correct. Hence a segment attempt costs $kg_t + c_s$ in expectation and succeeds with probability $(1 - q_t)^k$; geometric repetition and $n/k$ segments give the formula. For fixed $t$,

$$\log(T(t,k)/n) = \log(g_t + c_s/k) + \beta_t k,$$

whose derivative is $\beta_t - c_s/[k(g_t k + c_s)]$. Its unique zero gives the displayed quadratic solution. Integer enumeration proves the finite claim.

For a concrete witness, take $(\varepsilon, p_a, p_r, c_v, c_s) = (0.05, 0.01, 0.05, 0.2, 2)$ and $n = 1020$. No local check ($t = 0$) is minimised near $k = 5$ at cost $1.809n$; local checking alone needs $t = 2$ to make $nq_t \leq 0.05$ and costs $1.633n$. The hybrid $t = 1, k = 51$ has perfect final correctness and cost $1.407n$, strictly below both. If local plurality sampling is used instead, substitute its per-accepted-step cost and residual error for $(g_t, q_t)$; with only the Hoeffding error ceiling from Theorem 21 (Appendix D), the same expression is an upper bound rather than an equality. □

**Corollary 27** (Transfer-priced joint policy)**.** *Run the policy of Proposition 26 on the machine of Remark 9 (ram(1), $C = 2$, endpoint segment verifier under absorbing-state semantics, $k \mid n$) and charge every semantic block transfer $\lambda_{\text{blk}} \geq 0$ in addition to the weighted call cost. Then, pathwise, $Q_{\text{blk}} = n/k + V_s$ with $V_s$ the number of segment-verifier calls, and*

$$\mathbb{E}[\text{Cost}_\lambda] = \frac{n}{k}\Big[(kg_t + c_s + \lambda_{\text{blk}})(1 - q_t)^{-k} + \lambda_{\text{blk}}\Big].$$

*Proof.* Write $m = n/k$. The trace of Remark 9 writes the initial checkpoint once and each of the $m - 1$ accepted non-final checkpoints once, and every segment attempt, accepted or rejected, reads its checkpoint once; local checks see the two resident blocks (state, candidate), insert no block and move none. Hence $Q_{\text{blk}} = m + V_s$ pathwise, as in the remark's identity $Q_{\text{blk}} = n/k + N_{\text{ver}}$ for $t = 0$. A segment attempt succeeds with probability $(1 - q_t)^k$ independently of the others, so $\mathbb{E}[V_s] = m(1 - q_t)^{-k}$. Adding $\lambda_{\text{blk}}\mathbb{E}[Q_{\text{blk}}]$ to the cost $m(kg_t + c_s)(1 - q_t)^{-k}$ of Proposition 26 gives the display. The term $\lambda_{\text{blk}}m$ depends on $k$, so the optimum is not that of the proposition with $c_s$ replaced by $c_s + \lambda_{\text{blk}}$; it is found by enumerating the admissible $(t, k)$. □

**Proposition 28** (Conditional-hazard checkpoint envelope)**.** *Consider the regular length-k rollback policy with a perfect segment verifier of cost $c_s$. A segment attempt always executes all k transitions before verification and costs $k + c_s$. Drop independence, but retain the absorbing-state semantics and suppose that, for every prior run history and every position i whose preceding transitions in the current attempt were correct, the conditional error hazard lies in $\underline{\varepsilon} \le \Pr(E_i \mid E_1^c, \ldots, E_{i-1}^c, \mathscr{H}) \le \bar{\varepsilon} < 1$. The bounds must continue to hold after every rollback. Then, without independence within or between attempts, the expected cost for $k \mid n$ obeys*

$$\frac{n}{k}(k + c_s)(1 - \underline{\varepsilon})^{-k} \le \mathbb{E}[T(k)] \le \frac{n}{k}(k + c_s)(1 - \bar{\varepsilon})^{-k}.$$

*Thus the independent formula is recovered when the two hazards coincide. If only marginal error rates are known, neither this envelope nor sampling amplification follows in general: perfectly correlated replicas provide a counterexample.*

*Proof.* By the conditional chain rule, after any pre-attempt history $\mathscr{H}$ the probability that all $k$ transitions are correct lies in $[(1 - \bar{\varepsilon})^k, (1 - \underline{\varepsilon})^k]$. Let $A$ be the number of attempts until the perfect verifier accepts. Iterating the same conditional bounds after each failure sandwiches $\Pr(A > m \mid \mathscr{H})$ between the tails of geometric variables with those two success probabilities. Summing the tails gives $(1 - \underline{\varepsilon})^{-k} \le \mathbb{E}[A \mid \mathscr{H}] \le (1 - \bar{\varepsilon})^{-k}$. Multiplication by the fixed attempt cost and summation over $n/k$ segments proves the claim. Perfect correlation across replicas shows why a marginal error rate alone cannot imply a concentration gain. □

## F. Controlled execution checks

We use controlled simulators to check that the constructive schedules in Section 4–Section 6 respect the stated context and memory capacities and reproduce their analytical resource accounting. Communication, access, and recomputation checks use deterministic semantic transitions; reliability checks use stochastic transitions with prescribed error laws.

These experiments are implementation and scaling checks for the stated constructions. They are not treated as empirical evidence for the lower bounds themselves.

### F.1. Communication and access checks

**Bounded-context communication.** On grid DAGs, we compare row-major execution with a tiled schedule under the formal context bound. The tiled construction exhibits the $\Theta(n^2/C)$ communication scaling predicted by Corollary 11, whereas row-major execution remains quadratic until a live row fits in context. On sequential chains, full-history replay and dependency-only prompting reproduce the quadratic and linear input-accounting laws of Proposition 12.

**Memory access.** For pointer chasing, we compare random access with the demand-following, non-speculative sequential execution class of Theorem 2. Random access incurs one block read per pointer, while sequential access pays for the intervening head movement. The observed scaling is consistent with the separation proved in Theorem 2.

### F.2. Reliability checks

The reliability simulator implements the execution classes used in Section 7. Sampling-only policies commit stage by stage without correctness feedback, whereas verification and checkpoint policies receive the corresponding feedback and pay the stated verifier costs.

For checkpointing, simulated retry costs closely follow the exact objective $T(k)$ of Theorem 6, and the observed optimum tracks its closed-form minimizer. The square-root approximation is accurate only in its stated asymptotic regime; outside that regime the exact expression is the appropriate prediction.

### F.3. Recomputation implementation check

The simulator executes the constructive schedules from the proof of Theorem 3 while explicitly enforcing the context and persistent-memory capacities. Every model call requires its arguments to be resident, and an execution terminates on any context or storage overflow.

Across the tested capacity regimes, the checkpoint construction reproduces the predicted dependence

$$N = \left(\Theta\frac{n^2}{C + S} + n\right).$$

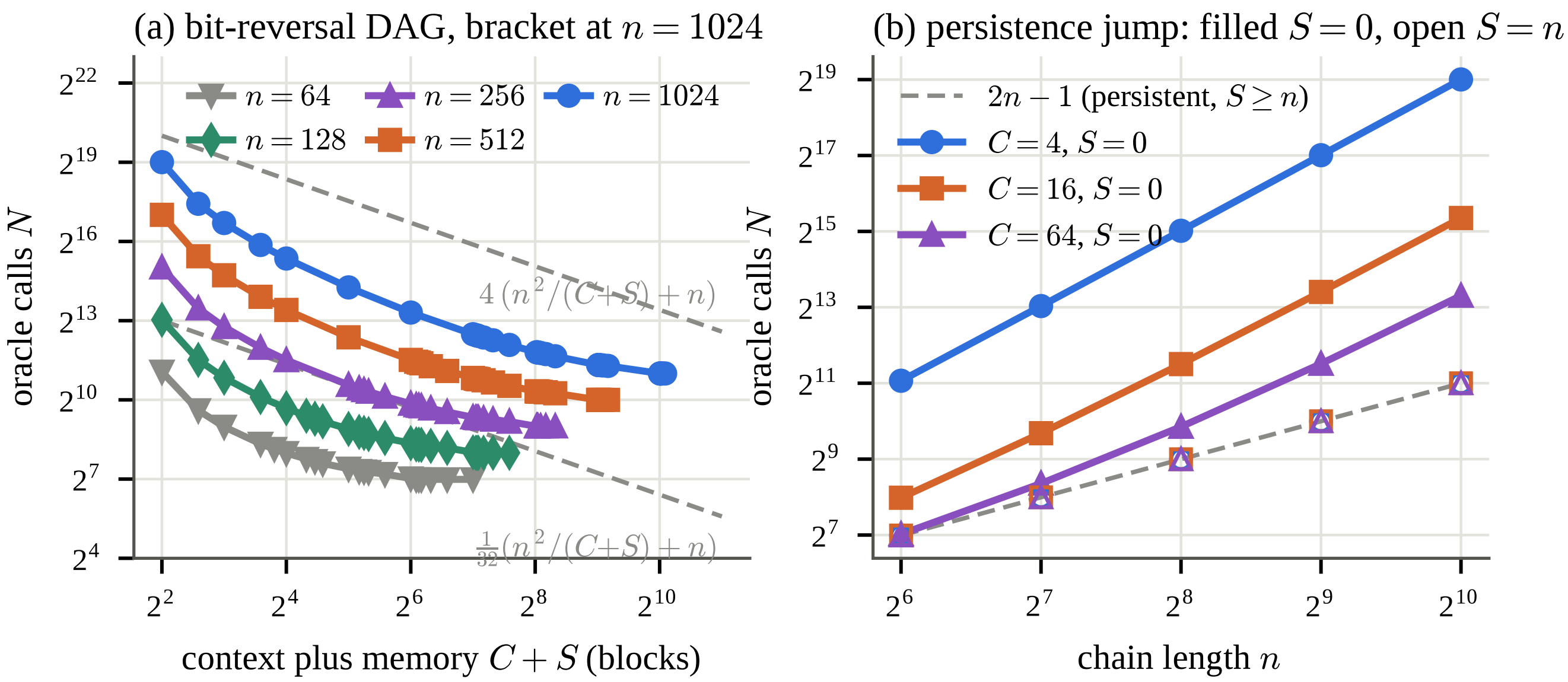

**Figure 3. Recomputation implementation check.** (a) Model-call count $N$ for the checkpoint construction of Theorem 3 on $\Delta_n^{\text{rev}}$ as total semantic capacity $C + S$ varies. Dashed curves show the theorem's bounds $\frac{1}{32}(n^2/(C+S)+n)$ and $4(n^2/(C+S)+n)$. (b) Call count as a function of $n$ without persistence ($S = 0$, filled) and with sufficient persistence ($S = n$, open); the dashed line $2n - 1$ is the persistent schedule.

Its measured schedule counts satisfy

$$\frac{N}{n^2/(C+S)+n} \in [0.58, 2.0],$$

within the explicit constants of Theorem 3, while its context–memory traffic satisfies $Q_{\text{blk}} \leq 2n$. With sufficient persistent storage, the construction uses exactly $2n - 1$ model calls, recovering linear call complexity (Figure 3).

The execution check also reproduces the feasibility boundary of the construction: at $C = 3$, no-persistence execution fails from insufficient semantic capacity, whereas adding one persistent block makes the schedule feasible. Thus the runs verify the capacity use and resource accounting of the constructive upper bounds; they are implementation checks rather than independent empirical evidence for the lower bound.

## G. GPT-6 Astra evaluation details

Prompts and evaluation artifacts are provided with the released code.

**Common protocol.** All experiments use GPT-6 Astra [1], accessed through the OpenAI API with reasoning effort set to `medium`, and a common harness. We record model and verifier calls, input and output traffic, cached input traffic, external-memory transfers, and peak prompt length. Calls producing semantic state are counted as model calls; verdict-only calls are counted as verifier calls.

The controller performs deterministic parsing, normalisation, and routing, but does not receive credit for semantic computation. Development or calibration data and held-out evaluation data are disjoint; forecast parameters and policy-selection rules determined from the former remain fixed on the latter. Success rates use Wilson 95% intervals, token means use chain-level bootstrap intervals, and paired comparisons use exact McNemar tests.

### G.1. Communication and cache accounting

**Synthetic chain.** We compare full-history replay, which repeatedly transmits the accumulated history, with dependency-only prompting, which supplies only the state required for the next transition. For fixed-width state and record encodings, their logical input traffic is

$$T_{\text{in}}^{\text{dep}} = nh, \qquad T_{\text{in}}^{\text{replay}} = nh + b\,\frac{n(n-1)}{2},$$

for fixed per-step overhead $h$ and history-record width $b$. The measured logical traffic follows these identities across the evaluated depths, recovering the linear-versus-quadratic separation of Proposition 12.

**Prefix caching.** Replay prompts share long prefixes, so logical traffic can overstate physical prefill when prefix caching is available. We therefore also evaluate an exact-prefix accounting. On the deepest synthetic setting, the replay-to-depen-dency-only ratio falls from 7.14 × in logical traffic to 1.72 × with exact-prefix caching; on chained MATH the corresponding ratio falls from 11.0 × to 3.05 ×. Thus caching narrows, but does not eliminate, the communication advantage of dependency-aware prompting.

### G.2. Reliability

**Calibration and controlled checks.** We evaluate reliability policies on modular-affine chains whose transitions are executed by the model. A disjoint calibration split estimates the candidate error rate together with the error rates and relative cost of the learned verifier. These quantities parameterize the sampling, local- verification, and checkpoint policies evaluated in Section 8.2. Controlled checkpoint experiments additionally use a perfect segment verifier at stipulated costs to isolate the prediction of Theorem 6 from verifier error.

**Dependence diagnostics.** Repeated sampling reveals substantial variation in error rates across chain states, so a pooled homogeneous i.i.d. model does not adequately describe the observed sampling process. The predeclared $m^\star = 15$ voting policy nevertheless solved 6/6 chains with 0/96 wrong plurality decisions.

Verifier errors are likewise candidate dependent: naturally wrong, correct, and synthetically corrupted candidates exhibit different acceptance behaviour. This observation motivates the candidate-dependent formulation of imperfect verification in Appendix D, rather than a single candidate-independent false-accept probability.

**Held-out checkpoint-policy tournament.** The tournament compares the principal reliability mechanisms studied in Section 7: unverified execution, sampling, local verification, segment rollback, and a joint local-verification–rollback policy. Calibration and held-out evaluation use disjoint chains. Forecast parameters and the policy-selection rule are determined from the calibration split and remain fixed on the held-out split. The resulting arm-level success and cost measurements are reported in Section 8.2 and Figure 2.

For segment rollback, the verifier checks the affine map obtained by composing the public transition constants within a segment. This composition is controller-side arithmetic rather than a charged semantic call. The tournament therefore tests prospective checkpoint-policy selection conditional on this implemented segment-check interface, rather than an end-to-end realization of the VPA restriction.

The sampling results additionally confirm that state heterogeneity materially affects reliability prediction: a model that preserves the observed state-dependent error structure describes held-out failures substantially better than a pooled homogeneous i.i.d. approximation.

### G.3. Bit-reversal implementation check

The real-model task instantiates $\Delta_n^{\mathrm{rev}}$ using fixed semantic vertex operations. Each model call receives only the blocks required by its scheduled vertex, and the controller enforces the prescribed context and persistent-memory capacities. We compare recomputing, persistent, and resident schedules together with a one-call Direct reference.

Rule-specific error and token costs are estimated on disjoint calibration data and held fixed when evaluating the scheduled executions. The schedules' model-call, recomputation, and context–memory transfer counts are determined analytically.

Across the evaluated instances, every prescribed schedule reproduces these analytical counts without a capacity violation. Persistent execution achieves linear model-call complexity with linear context–memory traffic, resident execution removes the transfer cost when all required state fits in context, and recomputing execution incurs the predicted additional semantic calls when persistent state is unavailable. Figure 2(d) summarizes these implementation checks.